\PassOptionsToPackage{table,svgnames}{xcolor}
\documentclass[11pt,letterpaper]{mystyle}
\usepackage{adjustbox,multirow,makecell,wrapfig}
\usepackage{algorithm,algorithmic}
\usepackage{listings,fontawesome5}
\usepackage[comma,authoryear,compress]{natbib}
\tcbuselibrary{listings,breakable,skins}
\usepackage{xurl}
\ifdefined\XeTeXversion
\fi

\newcommand{\MethodName}{\textit{\textcolor{darkgray}{\textbf{MotorMind}}}}
\newcommand{\bx}[1]{}
\newcommand{\huan}[1]{}
\newcommand{\realhuan}[1]{}

\title{\raisebox{-0.18em}{\includegraphics[height=1.15em]{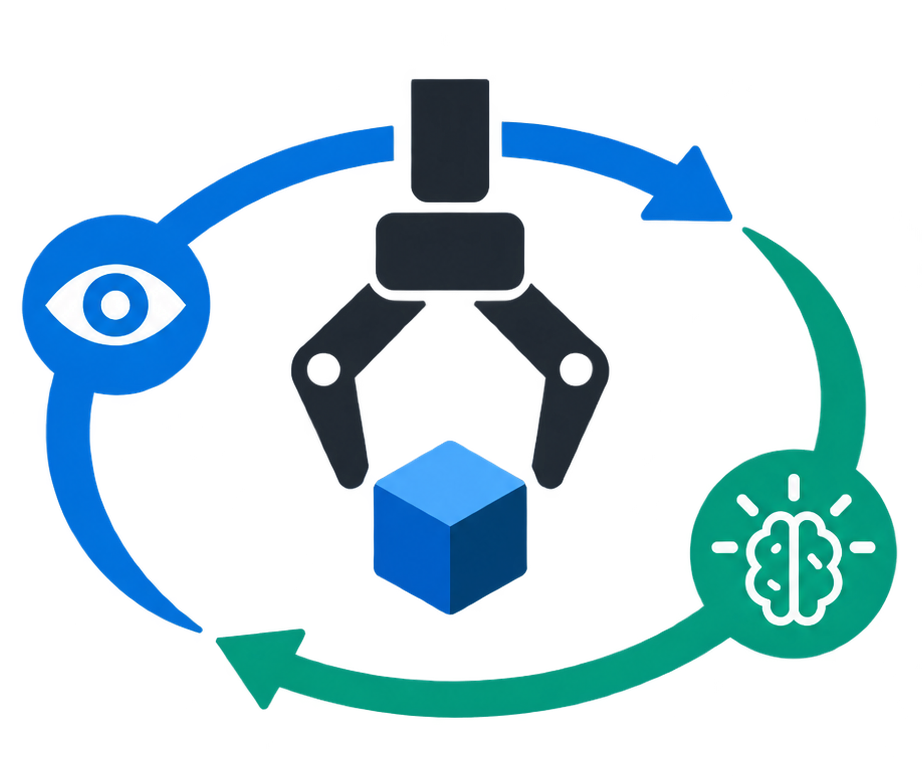}}\hspace{0.25em}MotorMind: Scaffolding General\\Vision Language Models for Zero-Shot Robot Manipulation}
\runningtitle{MotorMind: Scaffolding General Vision Language Models for Zero-Shot Robot Manipulation}

\author{Bingxuan Li\textsuperscript{*}}
\author{Siqi Song\textsuperscript{*}}
\author{Yizhuo Wu\textsuperscript{*}}
\author{Jiarui Yao}
\author{Tong Zhang}
\author{Huan Zhang}
\affil{\raisebox{-0.4em}{\includegraphics[height=1.3em]{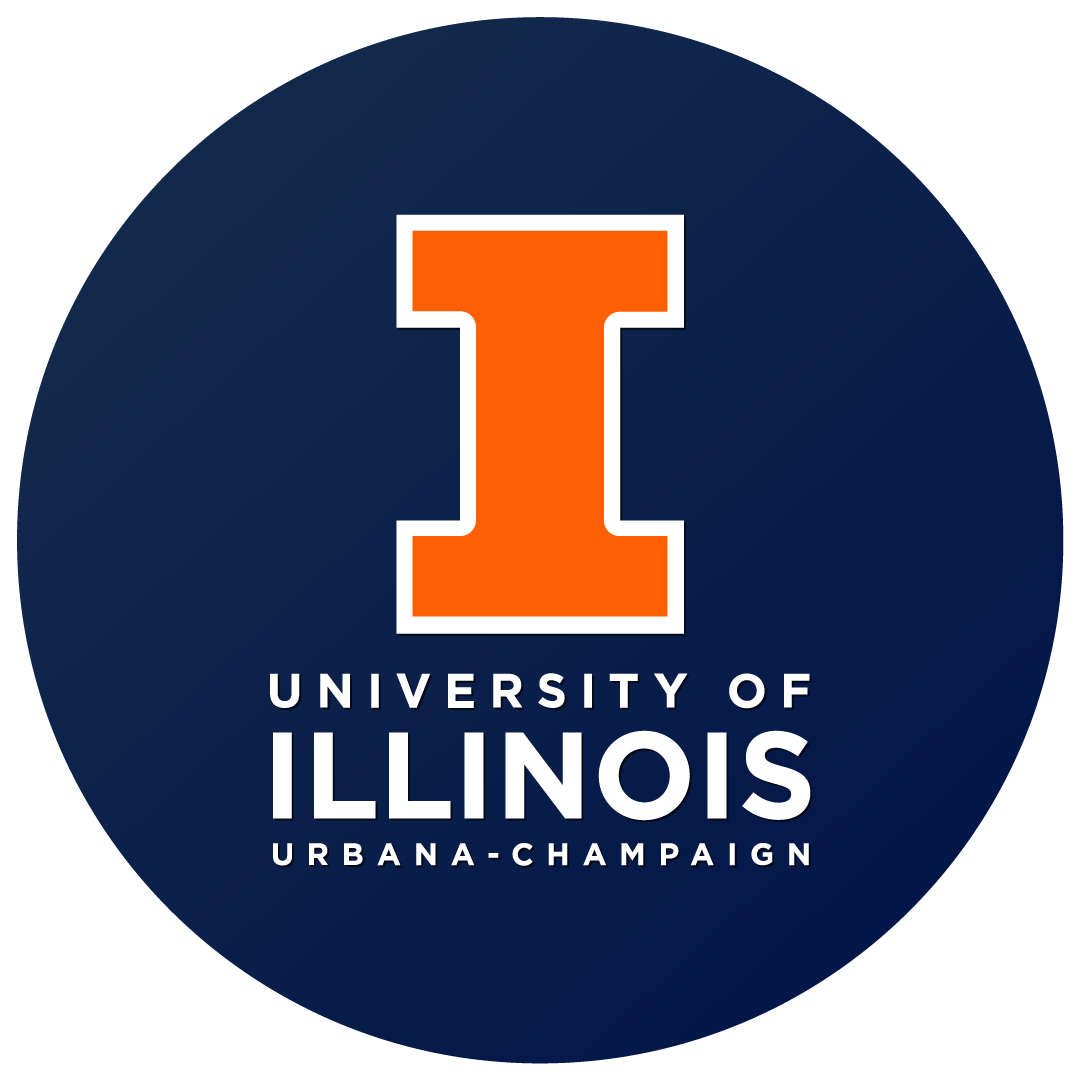}}\hspace{0.3em}University of Illinois Urbana-Champaign\\
\rule{0pt}{1.7em}{\small\textcolor{TinaCrimson}{\href{https://motor-mind.github.io}{\faGlobe\hspace{0.3em}https://motor-mind.github.io}}}}

\hypersetup{
  pdftitle={MotorMind: Scaffolding General Vision Language Models for Zero-Shot Robot Manipulation},
  pdfauthor={Bingxuan Li, Siqi Song, Yizhuo Wu, Jiarui Yao, Tong Zhang, Huan Zhang}
}
\addtocontents{toc}{\protect\setcounter{tocdepth}{-1}}
\begin{document}
\maketitle
\begingroup
\renewcommand{\thefootnote}{*}
\footnotetext[0]{Equal contribution, interchangeable order.}
\endgroup
\begingroup
\setlength{\intextsep}{6pt}
\begin{figure}[H]
    \captionsetup{font=footnotesize,skip=4pt}
    \centering
    \includegraphics[width=0.85\linewidth]{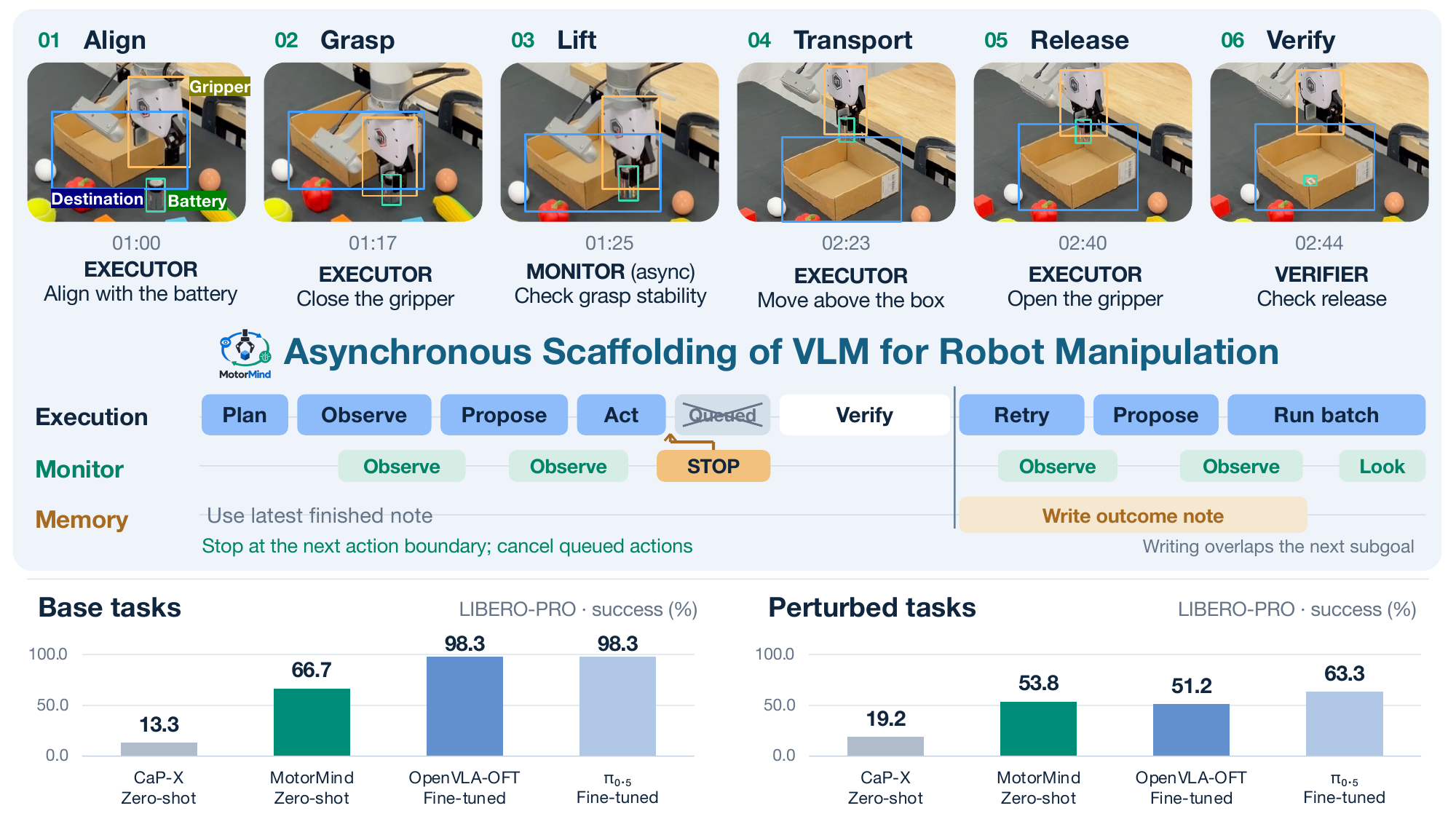}
    \caption{\textbf{\MethodName{}: Equipping general-purpose vision-language models into robot controller.} VLM proposes mid-level actions (e.g. "move forward by 12.20mm"), then pass to the control layer that connects those decisions to physical execution and feedback. Asynchronous monitoring enables interruption of pending actions, and outcome verification guides recovery and replanning. This design supports zero-shot manipulation across simulation and real-world settings without task-specific training.}
    \label{fig:teaser}
\end{figure}
\endgroup
\begingroup
\fontsize{10}{11.5}\selectfont
\setlength{\parskip}{3pt}
\titlespacing*{\section}{3pt}{7pt}{8pt}
\hypertarget{abstract}{}
\section*{Abstract}
Vision-language-action (VLA) models have advanced robotic manipulation, but their zero-shot generalization in new tasks and environments remains limited, and their reliance on specialized training keeps them from benefiting directly from rapidly advancing general-purpose vision-language models (VLMs). In parallel, recent agentic robotic systems leverage VLMs for high-level reasoning or coding agents for robot control, but often depend on extensive external models and tools, introducing additional complexity and cost. This motivates us to ask: \textbf{Can a general-purpose VLM itself operate a robot more like the human teleoperator by reasoning directly from observations, issuing actions, and continuously adapting to execution feedback, \textit{without} relying on external models such as learned action experts, coding agents or grounding tools like SAM3?} In this work, we introduce \MethodName{}, a robot manipulation harness that connects VLM-proposed mid-level actions to deterministic robot control and feedback, with asynchronous monitoring and background memory updates.Without task-specific policy training, coding agents, or additional grounding tools such as SAM3, \MethodName{} achieves 66.7\% success on the base LIBERO-PRO suites and 53.8\% under perturbations, compared with at most 13.3\% and 19.2\%, respectively, for the prior zero-shot methods we evaluate. The same interface reaches 95\% average success on a real xArm6 robot across direct manipulation and human-perturbation settings. Replacing the backbone with a stronger VLM further improves performance, while the remaining failures—primarily due to visual grounding, embodied reasoning, and action knowledge—decrease as VLM capability improves. These results show that a general-purpose VLM, when equipped with an appropriate mid-level action representation and asynchronous execution harness, can perform effective zero-shot robotic manipulation.
\par\endgroup

\clearpage








\section{Introduction}

Robotics foundation models such as vision-language-action (VLA) models have emerged as a promising paradigm for robotic manipulation by mapping visual observations and language instructions directly to robot actions~\citep{kim2025openvla, black2025pi05}.
Despite strong in-distribution performance, their zero-shot transfer remains limited under changes in object appearance, spatial configuration, task semantics, and environment structure.
In practice, strong performance in an unknown setting often relies on additional robot-action data or policy adaptation~\citep{kim2025fine}. In addition, these models were created via training on specialized robotics datasets and cannot directly benefit from the emerging intelligence from recent generic vision-language foundation models.

A parallel line of work explores \emph{agentic robotics}, where general-purpose VLMs are used for planning and high-level reasoning while external components handle perception, grounding, action generation, motion planning, or execution\citep{aspire2026, zhang2026harnessvla}. This avoids learning every stage of manipulation end-to-end, but often rely on substantial external models, including learned action experts\citep{black2025pi05}, segmentation model\citep{carion2025sam3segmentconcepts}, skill libraries\citep{aspire2026,fu2026capx}, motion planners\citep{huang2023voxposer}, and iterative execution modules\citep{chen2026etanewagenticparadigm}. As a result, advances in the spatial understanding and embodied reasoning capabilities of general-purpose VLMs have not necessarily been translated into simpler and more general robotic systems. This motivates a more direct question: 
\begin{center}
\textbf{Can a general-purpose VLM itself operate a robot more like the human teleoperator by reasoning directly from observations, issuing actions, and adapting to execution feedback, \textit{without} relying on external models such as learned action experts, coding agents or grounding tools like SAM3?}
\end{center}

To study this question, we evaluate three local decisions required during manipulation: selecting the next action, assessing progress toward a subgoal, and judging subgoal completion. Across the evaluated models, action selection is less accurate than either progress assessment or completion judgment, and even the stronger judgments remain imperfect. These findings identify limitations in local manipulation decisions; they do not establish that a particular control abstraction causes those errors.  Motivated by these findings, we introduce \MethodName{}, a VLM-centric manipulation harness designed around two principles. First, a compact mid-level action representation exposes parameterized translations, rotations, and gripper operations to the VLM, while embodiment-specific controllers convert these proposals into physical motion. Second, background monitoring checks updated observations during execution and can request cancellation of pending commands at the next action boundary. Motion proposals, execution, and outcome assessment form a sequential decision loop; memory summaries are generated asynchronously for subsequent planning and verification. This organization lets the system reconsider an invalid continuation without requiring each monitoring check to block execution.
This design enables the same VLM-facing reasoning and action interface to operate across different robot embodiments. In simulation, \MethodName{} achieves a 66.7\% success rate on the LIBERO-PRO base suite, compared with 13.3\% for the strongest prior method we evaluated without target-domain policy training\citep{fu2026capx}. Under semantic, object, position, and task perturbations, it achieves 53.8\% success, while the strongest prior zero-shot baseline reaches 19.2\%. Importantly, the same interface transfers to a \emph{real xArm6 robot without additional policy adaptation}, achieving 95\% average success across direct manipulation and human disturbance settings. These results show that a general-purpose VLM can provide a strong zero-shot decision-making core when coupled with an appropriate control abstraction and execution structure.

\MethodName{} also seamlessly improves with the VLM backbone: replacing the Qwen3.8-Flash-Next backbone (targeted for fast local robotics inference) with GPT-6 Sol (stronger but slower via API) further boosts the success rate on LIBERO-PRO base suite from 66.7\% to 83.3\%.
This suggests that \emph{improvements in general-purpose multimodal foundation models may translate directly into stronger robotic behavior} without requiring a redesign of the overall control architecture, or retraining with specialized data or model architecture.
Our extensive failure analysis sheds light on the error modes that stronger VLM models reduce,
with most failures arising from three sources: imperfect visual grounding, such as selecting the wrong object or target location; embodied reasoning errors, such as incorrectly determining that a task has completed; and insufficient action knowledge, where the model repeatedly proposes ineffective motions for particular situations.

\begin{table*}[t]
\centering
\scriptsize
\setlength{\tabcolsep}{2pt}
\renewcommand{\arraystretch}{1.12}

\begin{tabularx}{\textwidth}{
    @{}p{0.35\textwidth}
    @{\hspace{12pt}}
    p{0.2\textwidth}
    p{0.2\textwidth}
    >{\raggedright\arraybackslash}X@{}
}
\toprule
\textbf{Method} &
\textbf{\faRobot~Action Generation} &
\textbf{\faCubes~Extra Components} &
\textbf{\faBolt~Execution} \\
\midrule

$\pi_{0.5}$~\citep{black2025pi05} &
Learned action policy &
Trained action model &
Action chunks \\

MolmoAct2~\citep{molmoact2_2026} &
Learned action policy &
Trained action model&
Closed-loop prediction \\

OpenVLA-OFT~\citep{kim2025openvla} &
Trained action model &
Fine-tuned action decoder &
Action chunks \\

GR00T N1.5~\citep{gr00tn15} &
Trained action model &
Diffusion action model &
Policy rollout \\

\midrule

CaP-X~\citep{fu2026capx} &
Generated robot code &
Perception and control tools &
Program execution \\

VoLoAgent~\citep{chen2026volo} &
VLA + robot primitives &
Perception models and tools &
Tool orchestration \\

Harness VLA~\citep{zhang2026harnessvla} &
VLA + primitives &
VLA + primitive library &
Feedback-guided retries \\

\midrule

\textbf{\MethodName{}} &
\textbf{VLM Only} &
\textbf{None} &
\textbf{Asynchronous Execution} \\

\bottomrule
\end{tabularx}

\caption{
\textbf{Comparison of robotic manipulation methods.}
Most existing works rely on a learned action policy, generated robot code, or extra perception and control modules.
\MethodName{} uses the VLM directly for semantic action generation without
a coding agent, learned action model, SAM-based perception, or external inverse kinematics (IK) solver. See Appendix~\ref{app:related-work} for a detailed discussion of related work.
}
\label{tab:related_work_comparison}
\end{table*}


Overall, our results support a complementary direction to conventional VLA scaling. Rather than learning robotic behavior fully from demonstrations, increasingly capable VLMs can provide much of the semantic decision making required for manipulation when physical interaction is exposed at the right level of abstraction. This reframes zero-shot robotic control as a problem of \emph{aligning general-purpose model capabilities with control representation}, rather than solely learning specialized action policies.

\section{Diagnosing VLMs for Robotic Manipulation}
\label{sec:embodied-qa}

Using a general-purpose VLM as a robotic policy requires bridging semantic reasoning and embodiment-specific motion. We first examine how semantic decisions can be grounded into robot control and which VLM capabilities are sufficiently reliable for manipulation. The VLM policy must repeatedly answer three questions: \emph{What should happen next?} \emph{Did the previous action make progress?} \emph{Is the current subgoal complete?} We therefore study three capabilities:

\begin{itemize}[leftmargin=*, noitemsep, topsep=1pt]
    \item \textbf{Action Selection:} Choose the next action from the instruction and current observations.
    \item \textbf{Progress Assessment:} Determine whether an observed transition advances the task.
    \item \textbf{Subgoal Completion:} Determine whether the current state satisfies the stated subgoal and its success criterion.
\end{itemize}

\noindent\textbf{Diagnostic Benchmark.}
We construct an embodied question-answering benchmark with 240 questions, 80 for each capability. Action selection proposes future behavior, progress assessment interprets execution outcomes, and subgoal completion determines whether the active subgoal requires further actions. In the benchmark, action selection and progress assessment require the model to use past and current observations to predict the type of the next action and determine whether progress toward the current subgoal has been achieved. Subsubgoal completion requires the model to judge whether the stated criterion is satisfied from the current observation. These questions assess local visual decision making rather than closed-loop control of a complete task.
See Appendix~\ref{app:diagnostic} for more details.

\begin{figure*}[t]
    \centering
    \includegraphics[width=\textwidth]
    {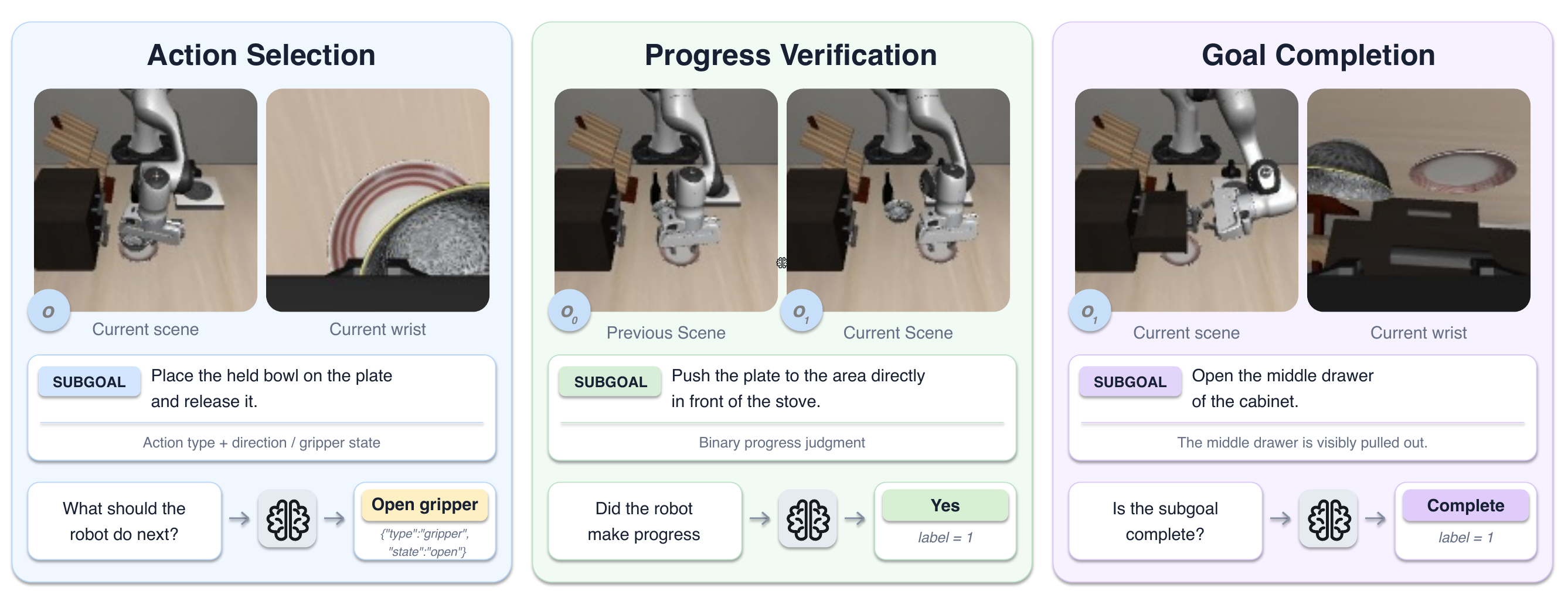}

    \caption{
    \textbf{Diagnosing VLM capabilities for robotic manipulation.}
    We evaluate action selection, progress assessment, and subgoal completion.
    }
    \label{fig:embodied_qa_overview}

\end{figure*}

\begin{table*}[t]
    \centering
    \scriptsize
    \setlength{\tabcolsep}{5.5pt}
    \renewcommand{\arraystretch}{1.05}

    \begin{tabular}{lccccc}
        \toprule
        \textbf{Model}
        & \textbf{Action}
        & \textbf{Progress}
        & \textbf{Completion}
        & \textbf{Overall}
        & \textbf{Latency $\downarrow$} \\
        \cmidrule(lr){2-5}
        \cmidrule(lr){6-6}

Qwen3.8-Flash-Next-FP8\citep{qwen3.8flashnext}
& 36.25
& \underline{55.00}
& \underline{65.00}
& \underline{52.08}
& \textbf{276} \\

HY-Embodied-0.5 MoT-2B\citep{x2026hyembodied05embodiedfoundationmodels}
& 20.00
& 50.00
& 47.50
& 39.17
& \underline{402} \\

Hy-Embodied-VLM-1.0 A3B\citep{wang2026hyembodiedvlm10efficientphysicalworldagents}
& 18.75
& 43.75
& 50.00
& 37.50
& 592 \\

Cosmos3-Nano\citep{nvidia2026cosmos3omnimodalworld}
& 18.75
& 50.00
& 62.50
& 43.75
& 3035 \\

GLM-5.3-Flash\citep{zai2026glm53flash}
& \underline{37.50}
& 52.50
& 58.75
& 49.58
& 403 \\

GPT-6 Astra\citep{openai2026gpt6astra}
& \textbf{60.00}
& \textbf{76.25}
& \textbf{82.50}
& \textbf{72.92}
& 8724 \\

        \bottomrule
    \end{tabular}

    \caption{
        \textbf{Diagnostic Results.}
        Accuracy (\%) on action selection, progress verification,
        subgoal completion, and all 240 questions overall.
        Latency denotes mean inference time per query in milliseconds.
    }
    \label{tab:embodied_qa}
\end{table*}

\noindent\textbf{Diagnostic Results and Analysis.}
Table~\ref{tab:embodied_qa} reveals three findings that inform the harness design. \textbf{First}, action selection is the least accurate capability for every evaluated model, ranging from 18.75\% to 60.00\%. For Qwen3.8-Flash-Next, action accuracy is 36.25\%, compared with 55.00\% for progress assessment and 65.00\% for completion. The answer spaces differ across these tasks, so these percentages describe performance on the diagnostic rather than equivalent measures of intrinsic difficulty. They motivate short action proposals that can be revised after observing their effects. \textbf{Second}, progress and completion judgments remain imperfect, motivating outcome assessment that combines model judgments with measured robot feedback. The diagnostic alone does not establish that repeated checks eliminate these errors. \textbf{Third}, GPT-6 Astra achieves the highest accuracy on all three capabilities but takes substantially longer per query: 8.724\,s, compared with 0.276\,s for Qwen3.8-Flash-Next. This accuracy--latency trade-off motivates considering both decision quality and feedback frequency when choosing the backbone. Together, the findings motivate a revisable, feedback-driven control interface; they do not isolate control abstraction as the cause of the observed errors.

\section{\MethodName{}: VLM Harness for Zero-Shot Robotic Manipulation}
\label{sec:method}

The analysis in Section~\ref{sec:embodied-qa} motivates three design choices for \MethodName{}. First, we use Qwen3.8-Flash-Next as a practical balance between manipulation reasoning capability and inference latency. Second, the system continuously refines short-horizon action decisions, allowing spatial errors to be corrected as interaction unfolds. Third, as VLMs are not fully reliable at assessing execution progress or task completion, \MethodName{} repeatedly checks progress and verifies subgoal outcomes. 

These choices lead to a harness that connects a frozen VLM to measured robot feedback through mid-level actions. Motion decisions follow a sequential execution loop, while monitoring and memory updates run asynchronously. 

\subsection{Task Formulation}
\label{sec:task-formulation}

Given a natural-language instruction $\ell$, the objective is to produce a sequence of robot actions that satisfies the requested physical outcome. At decision cycle $t$, the harness observes $o_t=(\mathcal{I}_t,r_t)$, where $\mathcal{I}_t$ contains the available camera images and $r_t$ contains measured robot state, including the tool pose and gripper state. \MethodName{} uses a frozen general-purpose VLM to make decisions from these observations without task-specific policy training.

To connect a task-level instruction to physical interaction, the Planner constructs an ordered sequence of subgoals $\mathcal{G}=(g_1,\ldots,g_M)$. Each subgoal specifies an intended state change, a target description, and a success criterion. For example, taking an object requires evidence that it is held, while carrying it requires both reaching the destination and retaining the grasp. The active subgoal therefore defines what the next actions should accomplish and what evidence is needed to accept their outcome. Because execution can invalidate the initial plan, the harness tracks which task requirements have been satisfied and can revise the remaining subgoals. The action interface below turns each active subgoal into a short, revisable sequence of motions.

\subsection{Mid-level Action Representation}
\label{sec:mid-level-actions}

Rather than asking the VLM to produce joint commands, \MethodName{} exposes parameterized actions in the robot's base frame. For each cycle, the Executor returns a structured proposal containing an assessment of the current subgoal, a completion signal, an optional action batch, and an expected outcome. A batch is represented as
\[
\mathcal{A}_t=(a_{t,1},\ldots,a_{t,K_t}), \qquad
 a_{t,k}=(\tau_{t,k},\boldsymbol{\eta}_{t,k}),
\]
where $\tau_{t,k}$ denotes the action type and $\boldsymbol{\eta}_{t,k}$ its parameters. The core manipulation primitives are \texttt{move}, \texttt{rotate}, and \texttt{gripper}. A translation specifies a direction or axis and a distance; a rotation specifies a direction or axis and an angle; a gripper action specifies opening or closing, optionally with a requested width. For example, a proposal can open the gripper, move left by a specified distance, and descend toward the target. The VLM selects the actions and their magnitudes, while the Controller converts them into embodiment-specific motion.

This interface gives the VLM control over how to pursue a subgoal while keeping the proposal interpretable and executable. The Controller validates each command, resolves its parameters into Cartesian motion, and records what the robot actually did. Subsequent proposals receive fresh observations, measured feedback, and recent execution history, so an intended movement is not treated as an achieved one. A completion signal requests outcome assessment; it does not by itself establish success. Appendix~\ref{app:motormind-grounding} describes the grounding and feedback mechanisms, and Appendix~\ref{app:motormind-actions} gives the remaining command fields and auxiliary actions. These representations provide the shared interface through which the roles below coordinate execution.

\subsection{Architecture Overview}
\label{sec:roles}
\label{sec:feedback}

\begin{figure}[h]
    \centering
    \includegraphics[width=0.95\linewidth]{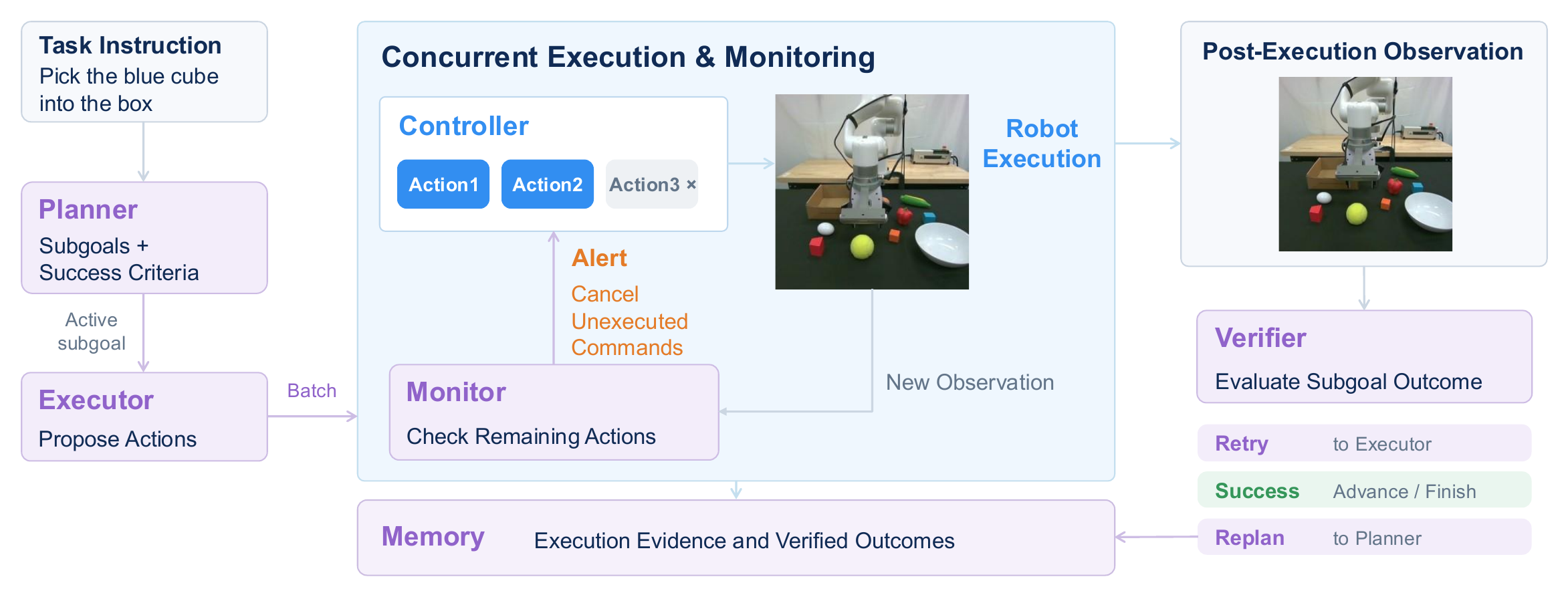}
    \caption{\textbf{Overview of \MethodName{}.} The Planner specifies subgoals and success criteria. The Executor proposes short action batches, which the Controller converts into robot motion. During execution, the Monitor evaluates updated observations and can request interruption at an action boundary, cancelling pending commands. Outcome assessment then determines whether to advance, retry, or replan. Memory summarizes execution evidence and assessed outcomes in the background for subsequent decisions.}
    \label{fig:method}
\end{figure}

\MethodName{} assigns five reasoning roles to the same general-purpose VLM: Planner, Executor, Monitor, Verifier, and Memory. The roles share model capabilities but receive different context and have different output authority. A deterministic Controller handles physical execution. Figure~\ref{fig:method} illustrates how a subgoal moves through these components; Appendix~\ref{app:motormind-vlm-roles-prompts} provides role-specific prompt excerpts.

The \textbf{Planner} translates the instruction into subgoals and revises the unfinished plan when execution evidence requires a change. For the active subgoal, the \textbf{Executor} uses current images, robot measurements, and recent cycle history to propose a short action batch. The \textbf{Controller} validates and executes the batch, checking later actions against fresh observations and returning measured motion and gripper outcomes. This closes the local decision loop: each new proposal is conditioned on the state reached by the previous attempt.

Two further roles evaluate whether that interaction remains consistent with the task. The \textbf{Monitor} watches the running subgoal and reports events such as a wrong target, a dropped object, or a changed scene. Its output is an alert, not a replacement action. At the end of an attempt, the \textbf{Verifier} assesses the success criterion using observations and robot-state evidence. Directly measured outcomes, such as a lost grasp or a measured release, can settle an attempt before a model verdict is needed. The resulting assessment determines whether the harness advances, retries the subgoal, or requests a revised plan; completing the plan also requires a task-level completion check.

Finally, \textbf{Memory} condenses execution evidence and assessed outcomes into a compact note for subsequent planning and verification. This note carries information across attempts, while the Executor's recent-cycle history supplies local context for motion decisions. The architecture thus separates action generation, ongoing monitoring, and outcome assessment. Its scheduling determines how these roles exchange evidence without making every background check block the robot.

\subsection{Asynchronous Scheduling}
\label{sec:async-loop}

\noindent\textbf{Sequential execution backbone.}
Within each subgoal, the main loop observes the scene, localizes the target when needed, constructs the Executor's context, requests a proposal, and executes its action batch. These operations remain sequential: a proposal depends on the current observation, and the next cycle depends on the measured outcome of the batch. Verification and replanning likewise follow the evidence that triggers them. Asynchrony is introduced around this decision loop, allowing monitoring and memory construction to proceed without waiting for the main loop to finish an entire subgoal.

\noindent\textbf{Concurrent monitoring and interruption.}
The Monitor runs on a background thread throughout a subgoal attempt, examining updated observations periodically and after motion steps. It evaluates whether the ongoing interaction remains consistent with the active subgoal and its success criterion. Informational alerts are recorded as evidence; a STOP alert requests cancellation of the running batch. The Controller reaches the next action boundary before stopping, and pending commands are discarded. For example, detecting that the robot has grasped the wrong object can prevent it from executing a queued placement action. The Monitor does not choose corrective motions: it interrupts a potentially invalid continuation and leaves the next decision to outcome assessment and recovery.

\noindent\textbf{Verification and recovery.}
Whether an attempt ends normally or through interruption, its outcome is assessed using updated observations, measured state, and any Monitor alert. An interruption is therefore a reason to reassess, not a verdict that the subgoal has failed or succeeded. If the criterion is satisfied, the harness advances; otherwise, it can retry the active subgoal or return the unresolved task requirements to the Planner. Monitoring for the ended attempt stops without waiting for an in-flight model call, and any late response from that attempt is discarded. The next attempt begins with its own monitoring context, preventing a stale alert from affecting a new subgoal.

\noindent\textbf{Non-blocking memory updates.}
After each outcome assessment, the harness requests a memory summary on a background writer. The next subgoal can begin while this summary is still being generated. Planning and verification read the latest completed note rather than waiting for the newest request, and pending requests are consolidated so the writer incorporates evidence accumulated since its last accepted summary. This preserves a sequential chain of motion decisions while allowing both scene monitoring and evidence summarization to overlap ongoing execution.

\section{Experiments}
\label{sec:experiments}

We evaluate whether the harness in Section~\ref{sec:method} turns a frozen general-purpose VLM into an effective manipulation system without task-specific policy training. Our experiments address two complementary questions: how well does \MethodName{} perform relative to zero-shot and task-fine-tuned baselines, and can it revise its decisions when the task changes during execution? We first evaluate task success, robustness to perturbations, and execution efficiency on LIBERO-PRO. We then test adaptation to moving objects, scene changes, and revised instructions, examining whether observation-driven execution remains effective when an initial plan is no longer sufficient.

\subsection{Main Experiment}

\noindent\textbf{Evaluation Settings.}
We evaluate \MethodName{} on the Goal, Spatial, and Object suites of
LIBERO-PRO\citep{zhou2026liberoprorobustfairevaluation}. The \emph{base evaluation} uses the original, unperturbed
episodes. The \emph{perturbation evaluation} measures performance under
Semantic, Object, Position, and Task perturbations. Together, these
settings test task completion in the original environments and
generalization when instructions, objects, layouts, or tasks change. See Appendix~\ref{app:simulation-evaluation} for more details.

\noindent\textbf{Baselines and Implementation.}
We compare with four VLA policies ($\pi_{0.5}$, MolmoAct2,
OpenVLA/OFT, and GR00T N1.5) in zero-shot and fine-tuned configurations where available. These VLA policy baselines directly predict low-level action tokens, providing a natural comparison for isolating the effect of our mid-level action representation. We also evaluate CaP-X and Harness VLA with their single-pass and iterative configurations, and VoLoAgent with either a fine-tuned or a zero-shot VLA backbone. These methods all follow sequential execution, providing a direct comparison for evaluating the effect of our asynchronous scheduling. We group methods according to whether their underlying action policy uses task-specific fine-tuning; agent-level test-time interaction alone does not place a method in the
fine-tuned group.\MethodName{} uses Qwen3.8-Flash-Next as its VLM backbone. It uses no LIBERO demonstrations, task-specific policy fine-tuning, or learned task-specific action model. Its VLM-facing semantic actions and asynchronous execution follow Sec.~\ref{sec:method}.

\noindent\textbf{Metrics.}
We report task success rate (SR, \%) and mean episode wall time in seconds as the primary metrics. As a secondary summary of their trade-off, we report a time-normalized success score,
\[
    \mathrm{TimeScore}=\frac{60\,s}{\bar{T}},
\]
where $s$ is the numerical success percentage on a 0--100 scale and $\bar{T}$ is the reported mean episode wall time in seconds for the same evaluation subset. The score is measured in percentage points per minute (pp/min); dividing it by 100 gives successes per minute when time is averaged over all evaluated episodes. For example, 66.7\% success in 223.4\,s gives 17.91\,pp/min. This is a descriptive time-normalized score, not a measure of monetary or compute cost. We interpret it alongside success and wall time, since short failed episodes do not indicate effective manipulation.



\definecolor{headerblue}{RGB}{31,48,108}

\newcommand{\best}[1]{\textbf{#1}}
\newcommand{\second}[1]{\underline{#1}}
\newcommand{\oursname}{\textcolor{headerblue}{\textbf{\MethodName}}}

\begin{table*}[!t]
\centering
\scriptsize
\setlength{\tabcolsep}{2.5pt}
\renewcommand{\arraystretch}{1.10}

\resizebox{\textwidth}{!}{%
\begin{tabular}{@{}ll*{13}{c}@{}}
\toprule
\multirow{2}{*}{\textbf{Method}}
& \multirow{2}{*}{\textbf{Configuration}}
& \multicolumn{4}{c}{\textcolor{headerblue}{\textbf{Base success rate (\%)}}}
& \multicolumn{5}{c}{\textcolor{headerblue}{\textbf{Perturbation success rate (\%)}}}
& \multicolumn{2}{c}{\textcolor{headerblue}{\textbf{Base efficiency}}}
& \multicolumn{2}{c}{\textcolor{headerblue}{\textbf{Perturbation efficiency}}}
\\
\cmidrule(lr){3-6}
\cmidrule(lr){7-11}
\cmidrule(lr){12-13}
\cmidrule(l){14-15}
&
& \textbf{Goal}
& \textbf{Spatial}
& \textbf{Object}
& \textbf{Avg.}
& \textbf{Semantic}
& \textbf{Object}
& \textbf{Position}
& \textbf{Task}
& \textbf{Avg.}
& \textbf{Time (s) $\downarrow$}
& \textbf{Time score $\uparrow$}
& \textbf{Time (s) $\downarrow$}
& \textbf{Time score $\uparrow$}
\\
\midrule

\multicolumn{15}{c}{%
\textcolor{headerblue}{\textbf{Fine-Tuned Policies / Agentic Methods with Fine-Tuned Policies}}}
\\
\midrule

$\pi_{0.5}$
& Fine-tuned
& \second{95.0} & \best{100.0} & \best{100.0} & \second{98.3}
& \best{98.3} & \best{95.0} & 36.7 & 23.3 & \second{63.3}
& \second{5.9} & \second{997.1} & \second{6.8} & \best{562.1}
\\

MolmoAct2
& Fine-tuned
& \best{100.0} & \best{100.0} & \best{100.0} & \best{100.0}
& 95.0 & \second{90.0} & 36.7 & 30.0 & 62.9
& 6.6 & 914.6 & 7.7 & \second{492.2}
\\

OpenVLA / OFT
& Fine-tuned
& \second{95.0} & \best{100.0} & \best{100.0} & \second{98.3}
& \second{96.7} & 86.7 & 11.7 & 10.0 & 51.2
& \best{5.7} & \best{1033.5} & \best{6.7} & 459.9
\\

GR00T N1.5
& Fine-tuned
& 0.0 & \second{95.0} & 0.0 & 31.7
& 30.0 & 30.0 & 1.7 & 16.7 & 19.6
& 15.6 & 121.8 & 16.2 & 72.4
\\

VoLoAgent (FT VLA as Tool)
& Zero-shot agent
& 55.0 & 90.0 & \best{100.0} & 81.7
& 88.3 & 83.3 & 45.0 & 25.0 & 60.4
& 106.2 & 46.1 & 216.9 & 16.7
\\

\multirow[c]{5}{*}{Harness VLA (FT VLA as Tool)}
& Zero-shot
& 35.0 & 30.0 & 30.0 & 31.7
& 30.0 & 31.7 & 28.3 & 25.0 & 28.8
& 354.6 & 5.4 & 342.5 & 5.0
\\
& S1: 5 loops
& 55.0 & 50.0 & 55.0 & 53.3
& 65.0 & 51.7 & \second{50.0} & \second{50.0} & 54.2
& 1040.4 & 3.1 & 1286.2 & 2.5
\\
& S1: 10 loops
& 70.0 & 90.0 & \second{80.0} & 80.0
& 75.0 & 66.7 & \best{53.3} & \best{66.7} & \best{65.4}
& 1764.0 & 2.7 & 2184.1 & 1.8
\\
& S2: 5 loops
& - & - & - & -
& 55.0 & 48.3 & 43.3 & 41.7 & 47.1
& - & - & 1125.2 & 2.5
\\
& S2: 10 loops
& - & - & - & -
& 70.0 & 58.3 & 41.7 & 45.0 & 53.8
& - & - & 1866.2 & 1.7
\\

\midrule
\multicolumn{15}{c}{\textcolor{headerblue}{\textbf{Zero-Shot Methods}}}
\\
\midrule

$\pi_{0.5}$
& Zero-shot
& 0.0 & 0.0 & 0.0 & 0.0
& 0.0 & 0.0 & 0.0 & 0.0 & 0.0
& \second{8.3} & 0.0 & \second{8.2} & 0.0
\\

MolmoAct2
& Zero-shot
& 0.0 & 0.0 & 0.0 & 0.0
& 0.0 & 0.0 & 0.0 & 0.0 & 0.0
& 9.4 & 0.0 & 9.3 & 0.0
\\

OpenVLA / OFT
& Zero-shot
& 0.0 & 0.0 & 0.0 & 0.0
& 0.0 & 0.0 & 0.0 & 3.3 & 0.8
& 34.7 & 0.0 & 34.4 & 1.5
\\

GR00T N1.5
& Zero-shot
& 0.0 & 0.0 & 0.0 & 0.0
& 0.0 & 0.0 & 0.0 & 0.0 & 0.0
& \best{6.3} & 0.0 & \best{6.2} & 0.0
\\

\multirow[c]{3}{*}{CaP-X (GPT5.6-Terra + Claude Opus 5)}
& Zero-shot
& 5.0 & 0.0 & 5.0 & 3.3
& 15.0 & 3.3 & 1.7 & 3.3 & 5.8
& 68.5 & \second{2.9} & 72.4 & \second{4.8}
\\
& 5 loops
& \second{25.0} & 0.0 & \second{15.0} & \second{13.3}
& 25.0 & \second{16.7} & \second{16.7} & 15.0 & 18.3
& 290.1 & 2.8 & 270.2 & 4.1
\\
& 10 loops
& \second{25.0} & 0.0 & \second{15.0} & \second{13.3}
& \second{26.7} & \second{16.7} & \second{16.7} & \second{16.7} & \second{19.2}
& 346.4 & 2.3 & 320.4 & 3.6
\\

VoLoAgent (zero-shot VLA as Tool)
& Zero-shot
& 0.0 & 0.0 & 0.0 & 0.0
& 0.0 & 0.0 & 0.0 & 0.0 & 0.0
& 500.0 & 0.0 & 500.0 & 0.0
\\

\midrule

\oursname{} (Qwen3.8-Flash-Next)
& Zero-shot
& \best{45.0} & \best{75.0} & \best{80.0} & \best{66.7}
& \best{58.3} & \best{46.7} & \best{51.7} & \best{58.3} & \best{53.8}
& 223.4 & \best{17.91} & 248.5 & \best{12.99}
\\

\bottomrule
\end{tabular}%
}

\caption{
\textbf{Main Evaluation Results.}
Success rates and averages are reported separately for unperturbed base tasks and
perturbed tasks; walltime and time-normalized success score use the corresponding evaluation
subset. Methods are grouped by whether their underlying policy uses task-specific
fine-tuning. Time score is $60s/\bar{T}$ in pp/min, where $s$ is success on a 0--100 scale and $\bar{T}$ is wall time in seconds; it does not measure monetary or compute cost.
}

\label{tab:combined_results}
\par\medskip
\begingroup
\captionsetup{type=figure}
    \centering
    \includegraphics[width=0.95\linewidth]{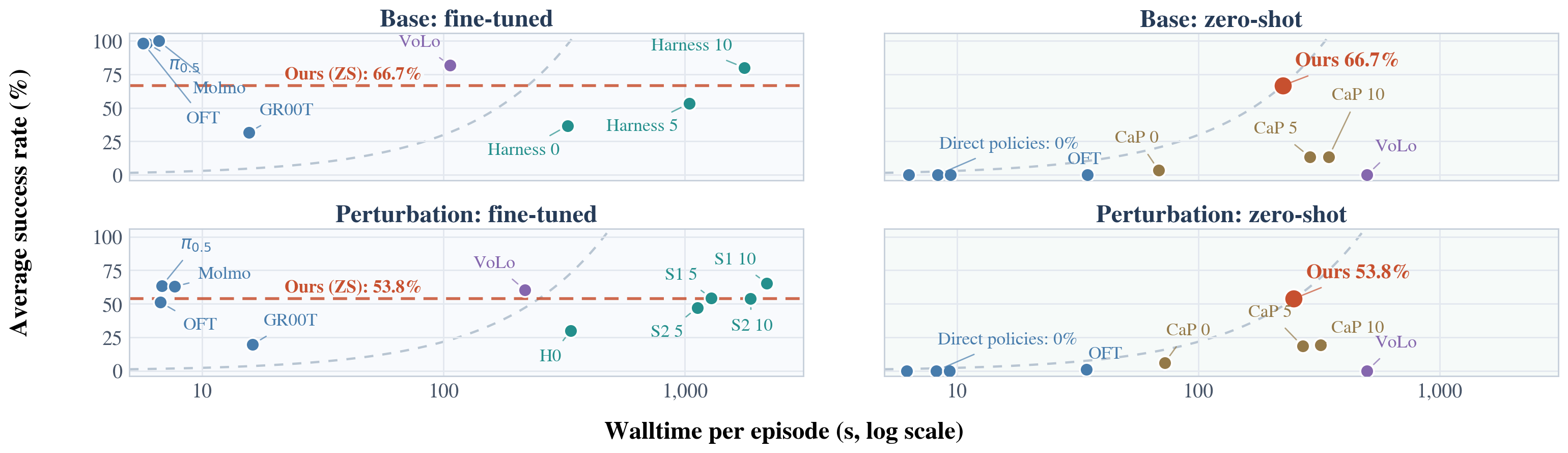}
    \caption{\textbf{Success rate versus episode wall time.}
    The four panels separate base and perturbation evaluations and
    distinguish methods with fine-tuned policies from zero-shot methods.
    Orange points denote \MethodName{}; orange horizontal lines show its
    success rate for reference in the fine-tuned panels. Gray dashed
    curves indicate the same time-normalized success score as \MethodName{}.}

    \label{fig:main_results_efficiency}
\endgroup
\end{table*}



\noindent\textbf{Experiment Results.}
\MethodName{} achieves the highest success rate among the
evaluated zero-shot methods, with average success rates of
66.7\% on the base suites and 53.8\% under perturbations
(Table~\ref{tab:combined_results}).
Without task-specific action-policy training, it achieves
75.0\% success on base Spatial tasks and 58.3\% under Task
perturbations.
As shown in Figure~\ref{fig:main_results_efficiency},
this zero-shot performance is also competitive with
task-fine-tuned policies: across 240 perturbed episodes,
\MethodName{} achieves 53.8\% average success, compared
with 51.2\% for OpenVLA-OFT.
The modest 2.6-percentage-point gap indicates comparable
performance, highlighting \MethodName{}'s ability to match
a task-fine-tuned policy without task-specific policy training.
These results demonstrate that \MethodName{} can translate
the capabilities of a pretrained multimodal model into
effective manipulation through observation-driven execution.
Its strong zero-shot performance across the evaluated suites
and perturbations supports this approach as a practical
alternative to task-specific policy fine-tuning.

\subsection{Adaptive Tasks Experiment}
We further test whether \MethodName{} can adapt its decisions to
changes that occur during execution. In the conveyor setting shown
on the right of Table~\ref{tab:dynamic_interactive_success}, objects
continue moving while the robot identifies and approaches its target,
requiring the agent to reason from the latest observations rather than
a static initial scene. We evaluate four groups of tasks:
\textbf{Dynamic Reasoning, Scene Shift, Dynamic Manipulation, and Prompt Shift}. Appendix~\ref{app:adaptive-reasoning} describes the task definitions, interventions, and evaluation protocol for all four groups.

\begin{itemize}[leftmargin=*, noitemsep, topsep=1pt]
    \item \textbf{Dynamic Reasoning:} Track and manipulate moving objects whose target is specified through semantic, relational, or temporal descriptions rather than named directly.
    \item \textbf{Scene Shift:} Keep the instruction fixed while moving the target object or destination during execution.
    \item \textbf{Dynamic Manipulation:} Track and manipulate objects moving on a conveyor during execution.
    \item \textbf{Prompt Shift:} Change the instruction after a physical trigger without resetting the robot or scene, requiring the agent to revise its decisions from the current state.
\end{itemize}

\begin{table*}[t]
\centering
\scriptsize

\begin{minipage}[c]{0.56\textwidth}
    \centering
    \setlength{\tabcolsep}{4.8pt}
    \renewcommand{\arraystretch}{1.12}

    \begin{tabular}{lcccc}
        \toprule
        \textbf{Method}
        & \shortstack{\textbf{Dynamic}\\
        \textbf{Reasoning}}
        & \shortstack{\textbf{Scene}\\
        \textbf{Shift}}
        & \shortstack{\textbf{Dynamic}\\\textbf{Manipulation}}
        & \shortstack{\textbf{Prompt}\\
        \textbf{Shift}} \\
        & \textit{(10 tasks)}
        & \textit{(10 tasks)}
        & \textit{(5 tasks)}
        & \textit{(5 tasks)} \\
        \cmidrule(lr){2-5}

        $\pi_{0.5}$     & 0\%  & \underline{70\%} & 0\%  & \underline{20\%} \\
        GR00T N1.5      & 0\%  & 10\%              & 0\%  & 0\% \\
        MolmoAct2       & 20\% & 50\% & \underline{40\%} & 0\% \\
        CaP-X           & \underline{30\%} & 20\%              & 20\% & \textbf{60\%} \\

        \midrule
        \textbf{\MethodName{}}
        & \textbf{70\%}
        & \textbf{90\%}
        & \textbf{80\%}
        & \textbf{60\%} \\

        \bottomrule
    \end{tabular}
\end{minipage}
\hspace{0.015\textwidth}
\begin{minipage}[c]{0.40\textwidth}
    \centering
    \includegraphics[
        width=\linewidth,
        trim=4 4 4 4,
        clip
    ]{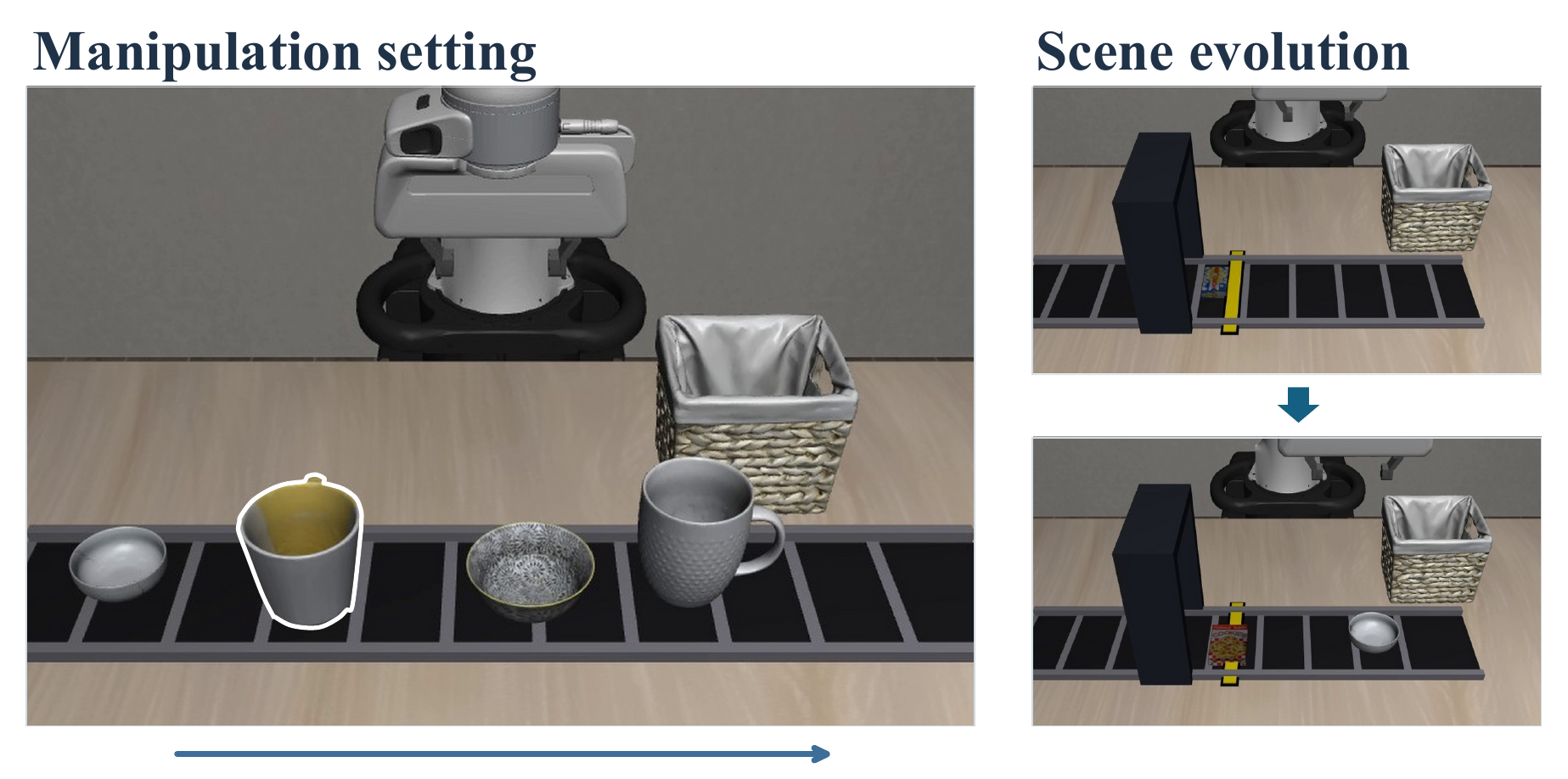}
\end{minipage}

\caption{
\textbf{Adaptive reasoning evaluation.}
\textbf{Left:} Success rates across four task groups.
\textbf{Right:} Example in which the robot must reason and act while the scene evolves.
Parentheses indicate the number of tasks; bold and underlined entries denote the best and second-best results, respectively.
}

\label{tab:dynamic_interactive_success}
\end{table*}

As shown in Table~\ref{tab:dynamic_interactive_success},
\MethodName{} achieves 70\% success on Dynamic Reasoning, 90\% on Scene Shift, 80\% on Dynamic Manipulation, and 60\% on Prompt Shift. The strongest baseline in each group achieves 30\%, 70\%, 40\%, and 60\%, respectively. These results suggest that
the same semantic action interface and asynchronous inference process
can support decisions even when the environment changes during execution.

\section{Real Robot Deployment}
\label{sec:real_robot}

We deploy \MethodName{} on a physical xArm6 robot to evaluate its
real-world performance with table-top tasks.
\MethodName{} receives observations from
the robot and RealSense D455 cameras and executes actions through
the standard xArm6 SDK, without task-specific demonstrations or
fine-tuning.
We design tasks in three settings to evaluate \MethodName{}'s
ability to perceive objects, adapt to scene changes, and
interpret instructions, respectively.

\noindent\textbf{Direct Perception.}
The instruction explicitly identifies both the object and its
destination, i.e., \textit{``Put the blue cube in the white
bowl.''} This setting tests whether the system can locate the
specified object and destination and execute the actions required
to complete the placement on the physical robot.

\begin{table}[H]
    \centering

    \begin{minipage}[c]{0.54\textwidth}
        \raggedleft
        \scriptsize
        \setlength{\tabcolsep}{4.5pt}
        \renewcommand{\arraystretch}{1.12}

        \begin{tabular}{lccccc}
            \toprule
            \textbf{Object}
            & \multicolumn{2}{c}{\textbf{Direct Perception}}
            & \multicolumn{2}{c}{\textbf{Human Intervention}}
            & \textbf{Avg. SR} \\
            \cmidrule(lr){2-3}
            \cmidrule(lr){4-5}
            &
            \textbf{Bowl} & \textbf{Box}
            & \textbf{Bowl} & \textbf{Box}
            & \\
            \midrule

            Blue Cube
            & 100\% & 100\%
            & 100\% & 100\%
            & 100\% \\

            Corn
            & 100\% & 100\%
            & 100\% & 80\%
            & 95\% \\

            Battery
            & 90\% & 100\%
            & 80\% & 90\%
            & 90\% \\

            \midrule
            \textbf{Avg. SR}
            & \textbf{97\%} & \textbf{100\%}
            & \textbf{93\%} & \textbf{90\%}
            & \textbf{95\%} \\
            \bottomrule
        \end{tabular}
    \end{minipage}%
    \hspace{0.025\textwidth}%
    \begin{minipage}[c]{0.36\textwidth}
        \centering
        \includegraphics[width=0.95\linewidth]{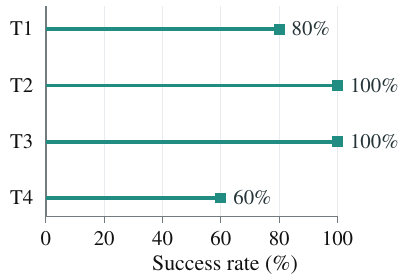}
    \end{minipage}

    \caption{
        \textbf{Real-robot zero-shot deployment.}
        \textbf{Left:} Success rates in Direct Perception and Human Perturbation tasks.
        \textbf{Right:} Success rates on Semantic Understanding tasks.
        Instructions include semantic categories, visual attributes, or spatial relations.
    }

    \label{tab:real_robot}
\par\medskip
\begingroup
\captionsetup{type=figure}
    \centering
    \includegraphics[width=0.92\linewidth]{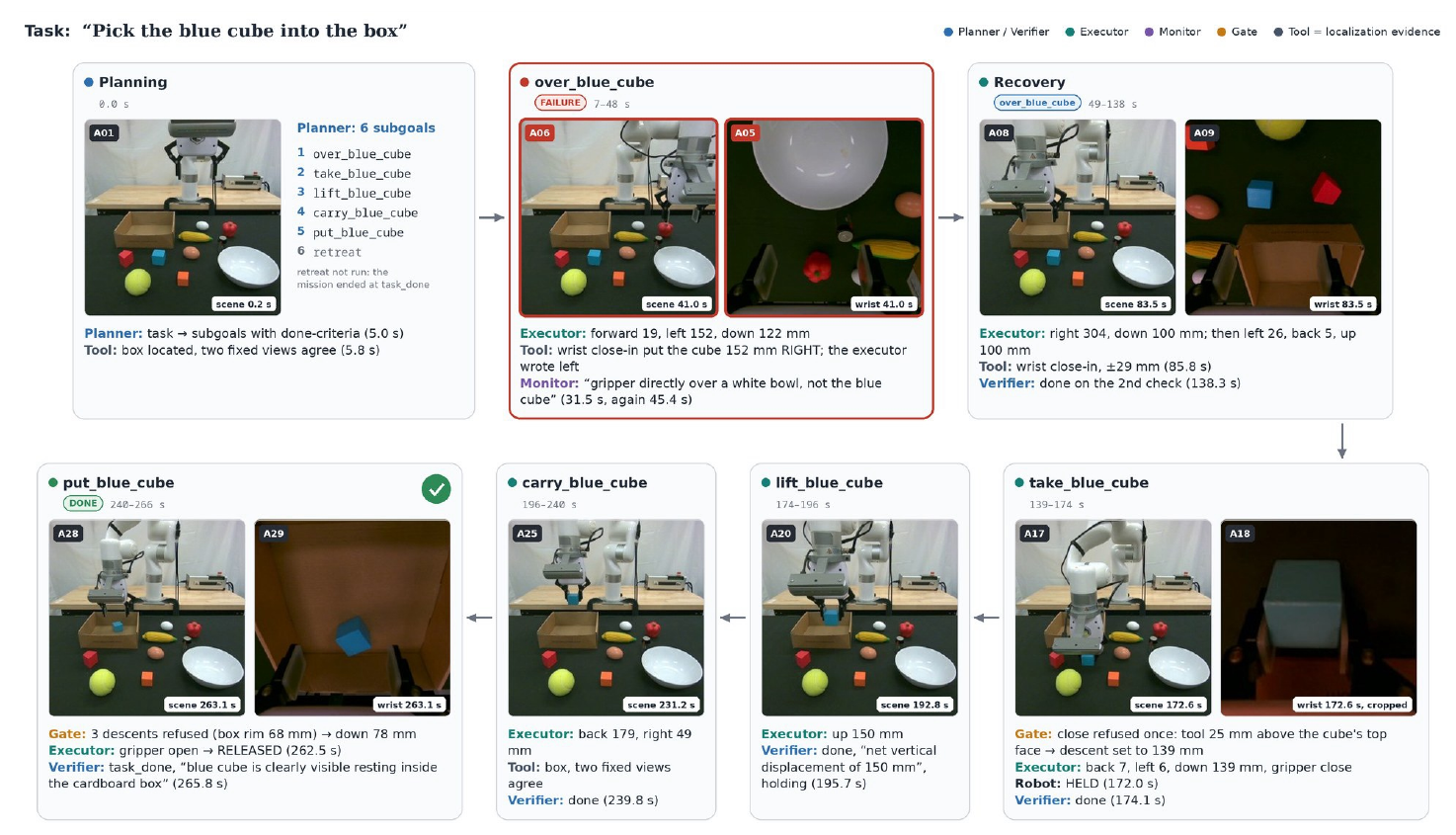}

    \caption{\textbf{Real-robot execution and recovery.}
    In this case, the instruction is to place the blue cube in the box,
    \MethodName{} initially moves toward an incorrect region near
    the white bowl. The Monitor identifies a mismatch with the
    intended subgoal. From a new observation, the Executor
    re-localizes the cube, grasps it, and completes the placement.}
    \label{fig:real_robot_case}
\endgroup
\end{table}

\begin{figure}[!t]
    \centering
    \includegraphics[width=0.95\linewidth]{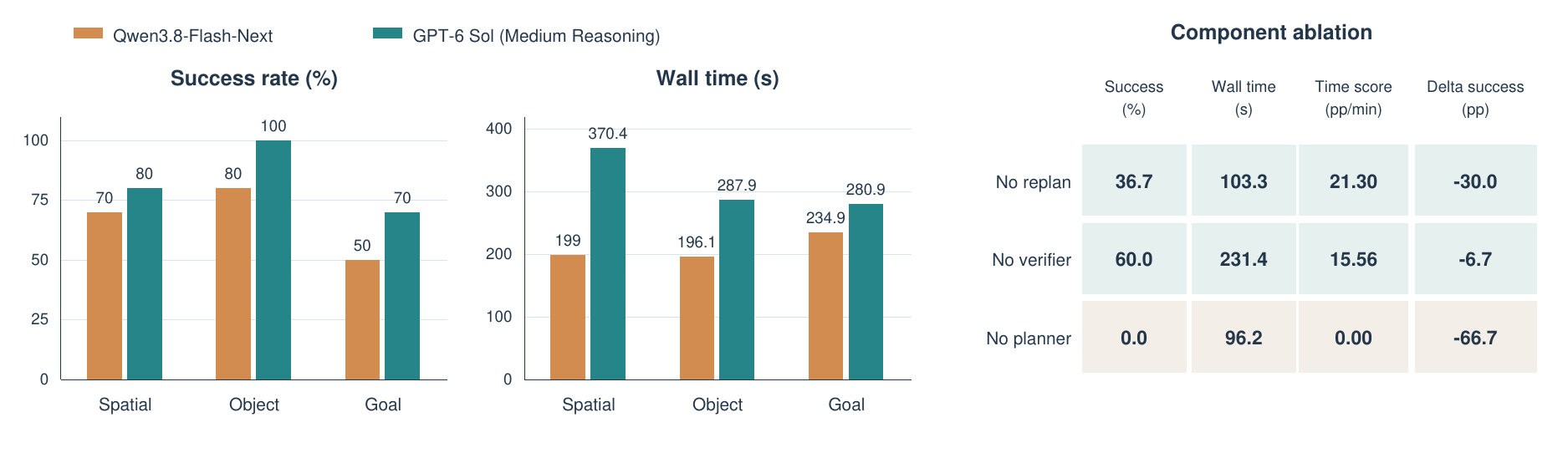}
    \caption{\textbf{Backbone sensitivity and component ablations.}
    Left: success rate and wall time across three LIBERO-PRO base task suites on single seed with
    Qwen3.8-Flash-Next and GPT-6 Sol (Medium Reasoning)
    Right: performance after removing replanning, verification, or planning.
    $\Delta$ Success is measured in percentage points relative to the full
    system's 66.7\% success rate. Time score uses the definition in Section~\ref{sec:experiments} (pp/min).}
    \label{fig:discussion_results}
\end{figure}

\noindent\textbf{Human Perturbation.}
We use the same object-destination tasks as in Direct Perception,
but introduce human interventions after execution begins.
A person deliberately moves or replaces the relevant object
during execution to disrupt task progress.
This setting tests whether the system can use subsequent
observations to detect the scene change and adapt its remaining
actions.

\noindent\textbf{Semantic Understanding.}
The instruction identifies the intended object or destination
through semantic categories (e.g., \textit{``the food item''}),
visual attributes (e.g., \textit{``the same color''}),
or spatial relations (e.g., \textit{``in the middle''}),
rather than explicitly naming it.
This setting tests whether the system can interpret the
instruction in the observed scene and execute
the corresponding action.

\paragraph{Results and Case Study}
In our evaluation,
each object-destination pair in Direct Perception
and Human Perturbations is evaluated over 10 trials per setting,
whereas each of the four Semantic Understanding tasks is
evaluated over 5 trials.
Object arrangements are randomized.

\noindent\textbf{Results.}
As shown in Table~\ref{tab:real_robot},
\MethodName{} achieves a pooled success rate of 95\% across Direct Perception (98\%) and Human Perturbations (92\%),
showing high placement success even under human interventions.
On the four Semantic Understanding tasks, success rates
are 80\%, 100\%, 100\%, and 60\% .
These results highlight \MethodName{}'s ability to translate
semantic descriptions into successful physical actions
across varied instruction types.

\noindent\textbf{Qualitative Case Study.}
Figure~\ref{fig:real_robot_case} illustrates how \MethodName{}
detects and corrects an execution error.
After the Planner decomposes the instruction into subgoals,
the initial approach positions the gripper above the white
bowl rather than the blue cube.
The Monitor flags this mismatch, and the Executor uses
updated visual observations to correct the gripper's position
before proceeding, and continue the task through
placement in the box. This rollout illustrates the role of
observation-driven correction during physical execution.

\section{Discussion}
\label{sec:discussion}

\begin{figure}
    \centering
    \includegraphics[width=1\linewidth]{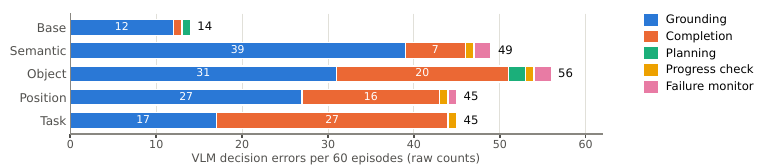}
    \caption{\textbf{Failure analysis of VLM decision errors across LIBERO-PRO settings.}}
    \label{fig:Failure}
\end{figure}

\paragraph{Backbone Sensitivity}
We replace Qwen3.8-Flash-Next with GPT-6 Sol (Medium Reasoning). As shown in Figure~\ref{fig:discussion_results}, GPT-6 Sol
improves success from 70\% to 80\% on Spatial, 80\% to 100\% on Object, and
50\% to 70\% on Goal. Its average success rate across these suites is 83.3\%,
compared with 66.7\% for Qwen3.8-Flash-Next. This gain comes with longer wall
time in every suite, most notably on Spatial (370.4\,s versus 199.0\,s).
These results suggest that the harness can benefit from a stronger reasoning
backbone. \bx{add a little bit more discussion here}
\huan{\textbf{Make the numbers agree with Table~\ref{tab:combined_results}.} The table gives Qwen Goal/Spatial/Object = 45/75/80\%; this section gives 50/70/80\%. Both average 66.7\%, which hides the mismatch. Wall time also differs (223.4 s in the table versus about 210 s here and in the appendix). Check which episodes and configuration each number comes from and use one set. Say whether both backbones ran on the same episodes. GPT-6 Sol here is a different model from GPT-6 Astra in the diagnostic, so do not link the two. Call this "backbone sensitivity", not a scaling law. For the \bx{} note, one useful sentence is: the gain is largest on Goal (+20 points) and the cost is wall time (370.4 s versus 199.0 s on Spatial).}

\paragraph{Ablation Study}
The ablations identify which reasoning functions sustain task success. Removing
replanning reduces success from 66.7\% to 36.7\%, a 30.0 percentage-point drop:
the system has fewer opportunities to recover when execution diverges from its
plan. Removing the verifier yields 60.0\% success, a smaller but measurable
6.7-point drop, indicating that explicit progress checks improve reliability.
Removing the planner is most damaging, reducing success to 0.0\%. The shorter
wall times of some ablated variants therefore do not indicate more effective
execution; in particular, the zero time-normalized success score without the planner must be
read alongside its zero success rate. Appendix~\ref{app:ablation} further examines the effect of the execution budget on success and wall time.
\huan{\textbf{Explain what each ablated variant still does.} "No planner" gives 0\%; say what the Executor receives in that case (the raw instruction? a completion criterion?) and whether it produced valid actions at all. If it produced none, the 0\% shows a broken setup rather than that planning is essential. Say what triggers replanning in the full system and what remains when the Verifier is removed (the evaluator's success check still ends the episode). The 6.7-point Verifier effect is 4 episodes out of 60; call it "small". Since a synchronous version was not run, say plainly that the ablations test the roles, not the schedule.}

\paragraph{Failure Analysis}
\label{sec:failure-analysis}

We categorize errors except malformed-output into grounding, completion, planning, progress checking, and failure monitoring. Grounding errors account for the largest share overall, particularly under Semantic, Object, and Position perturbations, while premature completion claims are also a major source of failure, especially for Task and Object perturbations. These results suggest that as foundation models continue to improve, a lightweight harness around a general-purpose VLM may become an increasingly practical path toward robust, general-purpose robotic manipulation. See Appendix~\ref{app:error-analysis} for more details.

\section{Conclusion}
We presented \MethodName{}, a zero-shot robotic manipulation harness that uses a single frozen VLM across five complementary roles, connects its reasoning to robot control through a mid-level action representation, and coordinates a sequential execution loop with concurrent monitoring, background memory updates, and explicit outcome verification. On LIBERO-PRO, \MethodName{} achieves 66.7\% success on the base suites and 53.8\% under perturbations, the highest among the zero-shot methods evaluated; it further reaches 95\% success on real-world xArm6 placement tasks, handles adaptive manipulation scenarios with moving objects, and improves to 83.3\% average success when using GPT-6 Sol as a stronger backbone. Our failure analysis shows that the remaining errors are dominated by inaccurate visual grounding and premature claims of task completion, suggesting that these bottlenecks can increasingly be addressed by stronger future VLMs with improved visual understanding and self-verification.

\clearpage

\bibliography{main}

\begin{thebibliography}{32}
\providecommand{\natexlab}[1]{#1}
\providecommand{\url}[1]{\texttt{#1}}
\expandafter\ifx\csname urlstyle\endcsname\relax
  \providecommand{\doi}[1]{doi: #1}\else
  \providecommand{\doi}{doi: \begingroup \urlstyle{rm}\Url}\fi

\bibitem[Black et~al.(2025{\natexlab{a}})Black, Brown, Darpinian, Dhabalia, Driess, Esmail, Equi, Finn, Fusai, Galliker, Ghosh, Groom, Hausman, Ichter, Jakubczak, Jones, Ke, LeBlanc, Levine, Li-Bell, Mothukuri, Nair, Pertsch, Ren, Shi, Smith, Springenberg, Stachowicz, Tanner, Vuong, Walke, Walling, Wang, Yu, and Zhilinsky]{black2025pi05}
Kevin Black, Noah Brown, James Darpinian, Karan Dhabalia, Danny Driess, Adnan Esmail, Michael~Robert Equi, Chelsea Finn, Niccolo Fusai, Manuel~Y. Galliker, Dibya Ghosh, Lachy Groom, Karol Hausman, Brian Ichter, Szymon Jakubczak, Tim Jones, Liyiming Ke, Devin LeBlanc, Sergey Levine, Adrian Li-Bell, Mohith Mothukuri, Suraj Nair, Karl Pertsch, Allen~Z. Ren, Lucy~Xiaoyang Shi, Laura Smith, Jost~Tobias Springenberg, Kyle Stachowicz, James Tanner, Quan Vuong, Homer Walke, Anna Walling, Haohuan Wang, Lili Yu, and Ury Zhilinsky.
\newblock {$\pi_{0.5}$}: A vision-language-action model with open-world generalization.
\newblock In \emph{Proceedings of The 9th Conference on Robot Learning (CoRL)}, volume 305 of \emph{Proceedings of Machine Learning Research}, pages 17--40. PMLR, 2025{\natexlab{a}}.
\newblock URL \url{https://proceedings.mlr.press/v305/black25a.html}.

\bibitem[Black et~al.(2025{\natexlab{b}})Black, Brown, Driess, Esmail, Equi, Finn, Fusai, Groom, Hausman, Ichter, Jakubczak, Jones, Ke, Levine, Li-Bell, Mothukuri, Nair, Pertsch, Shi, Smith, Tanner, Vuong, Walling, Wang, and Zhilinsky]{black2024pi0visionlanguageactionflowmodel}
Kevin Black, Noah Brown, Danny Driess, Adnan Esmail, Michael~Robert Equi, Chelsea Finn, Niccolo Fusai, Lachy Groom, Karol Hausman, Brian Ichter, Szymon Jakubczak, Tim Jones, Liyiming Ke, Sergey Levine, Adrian Li-Bell, Mohith Mothukuri, Suraj Nair, Karl Pertsch, Lucy~Xiaoyang Shi, Laura Smith, James Tanner, Quan Vuong, Anna Walling, Haohuan Wang, and Ury Zhilinsky.
\newblock $\pi_0$: A vision-language-action flow model for general robot control.
\newblock In \emph{Proceedings of Robotics: Science and Systems (RSS)}, 2025{\natexlab{b}}.
\newblock \doi{10.15607/RSS.2025.XXI.010}.
\newblock URL \url{https://www.roboticsproceedings.org/rss21/p010.html}.

\bibitem[Brohan et~al.(2023)Brohan, Brown, Carbajal, et~al.]{brohan2023rt2}
Anthony Brohan, Noah Brown, Justice Carbajal, et~al.
\newblock {RT-2}: Vision-language-action models transfer web knowledge to robotic control.
\newblock In \emph{Proceedings of The 7th Conference on Robot Learning (CoRL)}, volume 229 of \emph{Proceedings of Machine Learning Research}, pages 2165--2183. PMLR, 2023.
\newblock URL \url{https://proceedings.mlr.press/v229/zitkovich23a.html}.

\bibitem[Carion et~al.(2026)Carion, Gustafson, Hu, Debnath, Hu, Suris, Ryali, Alwala, Khedr, Huang, Lei, Ma, Guo, Kalla, Marks, Greer, Wang, Sun, Rädle, Afouras, Mavroudi, Xu, Wu, Zhou, Momeni, Hazra, Ding, Vaze, Porcher, Li, Li, Kamath, Cheng, Dollár, Ravi, Saenko, Zhang, and Feichtenhofer]{carion2025sam3segmentconcepts}
Nicolas Carion, Laura Gustafson, Yuan-Ting Hu, Shoubhik Debnath, Ronghang Hu, Didac Suris, Chaitanya Ryali, Kalyan~Vasudev Alwala, Haitham Khedr, Andrew Huang, Jie Lei, Tengyu Ma, Baishan Guo, Arpit Kalla, Markus Marks, Joseph Greer, Meng Wang, Peize Sun, Roman Rädle, Triantafyllos Afouras, Effrosyni Mavroudi, Katherine Xu, Tsung-Han Wu, Yu~Zhou, Liliane Momeni, Rishi Hazra, Shuangrui Ding, Sagar Vaze, Francois Porcher, Feng Li, Siyuan Li, Aishwarya Kamath, Ho~Kei Cheng, Piotr Dollár, Nikhila Ravi, Kate Saenko, Pengchuan Zhang, and Christoph Feichtenhofer.
\newblock {SAM 3}: Segment anything with concepts.
\newblock In \emph{International Conference on Learning Representations (ICLR)}, 2026.
\newblock URL \url{https://arxiv.org/abs/2511.16719}.

\bibitem[Chen et~al.(2026{\natexlab{a}})Chen, Hadfield, Zook, Uy, Song, Coumans, Yang, Ladhak, Qu, Birchfield, Tremblay, and Blukis]{chen2026volo}
Siyi Chen, Hugo Hadfield, Alex Zook, Mikaela~Angelina Uy, Chan~Hee Song, Erwin Coumans, Xuning Yang, Faisal Ladhak, Qing Qu, Stan Birchfield, Jonathan Tremblay, and Valts Blukis.
\newblock {VoLo}: A physical orchestrator for open-vocabulary long-horizon manipulation.
\newblock \emph{arXiv preprint arXiv:2606.07723}, 2026{\natexlab{a}}.
\newblock URL \url{https://arxiv.org/abs/2606.07723}.

\bibitem[Chen et~al.(2026{\natexlab{b}})Chen, Bai, Cao, Zeng, Lin, Lin, Liang, Ma, Huang, and Shou]{chen2026showharness}
Yanzhe Chen, Zechen Bai, Zhijun Cao, Wenzheng Zeng, Kevin~Qinghong Lin, Yiqi Lin, Guoqiang Liang, Kevin~Yuchen Ma, Qiming Huang, and Mike~Zheng Shou.
\newblock {Show-Harness}: Just a {VLM} agent can play robots.
\newblock \emph{arXiv preprint arXiv:2609.10522}, 2026{\natexlab{b}}.
\newblock URL \url{https://arxiv.org/abs/2609.10522}.

\bibitem[Chen et~al.(2026{\natexlab{c}})Chen, Huai, Li, Wang, Zhang, Zhang, Chen, Gong, Jiang, and Qiu]{chen2026etanewagenticparadigm}
Yitong Chen, Zezheng Huai, Sixian Li, Yubang Wang, Haozhe Zhang, Yifei Zhang, Hechang Chen, Jingjing Gong, Yu-Gang Jiang, and Xipeng Qiu.
\newblock {ETA}: A new agentic paradigm for embodied tasks.
\newblock \emph{arXiv preprint arXiv:2608.03924}, 2026{\natexlab{c}}.
\newblock URL \url{https://arxiv.org/abs/2608.03924}.

\bibitem[Elmaaroufi et~al.(2026)Elmaaroufi, Svegliato, Kalade, Schelle, Seshia, and Zaharia]{elmaaroufi2026rho}
Karim Elmaaroufi, Justin Svegliato, Sarunas Kalade, Graham Schelle, Sanjit~A. Seshia, and Matei Zaharia.
\newblock {RHO}: Your coding agent is secretly a roboticist.
\newblock \emph{arXiv preprint arXiv:2606.16458}, 2026.
\newblock URL \url{https://arxiv.org/abs/2606.16458}.

\bibitem[Fang et~al.(2026)Fang, Duan, Clay, Wang, Liu, Huang, Fan, Tsai, Chen, Wang, Xing, Cho, Park, Eftekhar, Sushko, Farley, Wadhwa, Harrison, Han, Lee, VanderBilt, Hendrix, Ellawela, Ngoo, Chai, Ren, Farhadi, Fox, and Krishna]{molmoact2_2026}
Haoquan Fang, Jiafei Duan, Donovan Clay, Sam Wang, Shuo Liu, Weikai Huang, Xiang Fan, Wei-Chuan Tsai, Shirui Chen, Yi~Ru Wang, Shanli Xing, Jaemin Cho, Jae~Sung Park, Ainaz Eftekhar, Peter Sushko, Karen Farley, Angad Wadhwa, Cole Harrison, Winson Han, Ying-Chun Lee, Eli VanderBilt, Rose Hendrix, Suveen Ellawela, Lucas Ngoo, Joyce Chai, Zhongzheng Ren, Ali Farhadi, Dieter Fox, and Ranjay Krishna.
\newblock {MolmoAct2}: Action reasoning models for real-world deployment.
\newblock \emph{arXiv preprint arXiv:2605.02881}, 2026.
\newblock URL \url{https://arxiv.org/abs/2605.02881}.

\bibitem[Fu et~al.(2026)Fu, Yu, El-Refai, Kou, Xue, Huang, Xiao, Li, Shi, Wu, Sastry, Zhu, Goldberg, and Fan]{fu2026capx}
Letian Fu, Justin Yu, Karim El-Refai, Ethan Kou, Haoru Xue, Huang Huang, Wenli Xiao, Fei-Fei Li, Guanya Shi, Jiajun Wu, Shankar Sastry, Yuke Zhu, Ken Goldberg, and Linxi Fan.
\newblock {CaP-X}: A framework for benchmarking and improving coding agents for robot manipulation.
\newblock In \emph{Proceedings of the 43rd International Conference on Machine Learning (ICML)}, 2026.
\newblock URL \url{https://arxiv.org/abs/2603.22435}.

\bibitem[Ghosh et~al.(2024)Ghosh, Walke, Pertsch, Black, Mees, Dasari, Hejna, Kreiman, Xu, Luo, Tan, Chen, Vuong, Xiao, Sanketi, Sadigh, Finn, and Levine]{octomodelteam2024octoopensourcegeneralistrobot}
Dibya Ghosh, Homer~Rich Walke, Karl Pertsch, Kevin Black, Oier Mees, Sudeep Dasari, Joey Hejna, Tobias Kreiman, Charles Xu, Jianlan Luo, You~Liang Tan, Lawrence~Yunliang Chen, Quan Vuong, Ted Xiao, Pannag~R. Sanketi, Dorsa Sadigh, Chelsea Finn, and Sergey Levine.
\newblock Octo: An open-source generalist robot policy.
\newblock In \emph{Robotics: Science and Systems (RSS)}, 2024.
\newblock \doi{10.15607/RSS.2024.XX.090}.
\newblock URL \url{https://arxiv.org/abs/2405.12213}.

\bibitem[Guo et~al.(2026)Guo, Mo, Wang, Zhang, Wang, Deng, Zhang, Rao, and Hu]{guo2026robodawn}
Meng-Hao Guo, Zhe-Han Mo, Jia-Jun Wang, Yi~Zhang, Kejin Wang, Yi-Xuan Deng, Jia-Peng Zhang, Yongming Rao, and Shi-Min Hu.
\newblock Transferring the intelligence of {VLMs} to robotic control.
\newblock \emph{arXiv preprint arXiv:2609.22966}, 2026.
\newblock URL \url{https://arxiv.org/abs/2609.22966}.

\bibitem[Hancock et~al.(2026)Hancock, Wu, Zha, Russakovsky, and Majumdar]{hancock2025actionslanguagefinetuningvlms}
Asher~J. Hancock, Xindi Wu, Lihan Zha, Olga Russakovsky, and Anirudha Majumdar.
\newblock Actions as language: Fine-tuning {VLMs} into {VLAs} without catastrophic forgetting.
\newblock In \emph{International Conference on Learning Representations (ICLR)}, 2026.
\newblock URL \url{https://openreview.net/forum?id=sFO9d6XSlf}.

\bibitem[Hu et~al.(2026)Hu, Sundaresan, Gao, and Sadigh]{hu2026via}
Hengyuan Hu, Priya Sundaresan, Jensen Gao, and Dorsa Sadigh.
\newblock {VIA}: Visual interface agent for robot control.
\newblock \emph{arXiv preprint arXiv:2607.11119}, 2026.
\newblock URL \url{https://arxiv.org/abs/2607.11119}.

\bibitem[Huang et~al.(2023)Huang, Wang, Zhang, Li, Wu, and Fei-Fei]{huang2023voxposer}
Wenlong Huang, Chen Wang, Ruohan Zhang, Yunzhu Li, Jiajun Wu, and Li~Fei-Fei.
\newblock {VoxPoser}: Composable {3D} value maps for robotic manipulation with language models.
\newblock In \emph{Proceedings of the 7th Conference on Robot Learning}, volume 229 of \emph{Proceedings of Machine Learning Research}, pages 540--562, 2023.
\newblock URL \url{https://proceedings.mlr.press/v229/huang23b.html}.

\bibitem[Huang et~al.(2025)Huang, Wang, Li, Zhang, and Fei-Fei]{huang2024rekep}
Wenlong Huang, Chen Wang, Yunzhu Li, Ruohan Zhang, and Li~Fei-Fei.
\newblock {ReKep}: Spatio-temporal reasoning of relational keypoint constraints for robotic manipulation.
\newblock In \emph{Proceedings of the 8th Conference on Robot Learning}, volume 270 of \emph{Proceedings of Machine Learning Research}, pages 4573--4602. PMLR, 2025.
\newblock URL \url{https://proceedings.mlr.press/v270/huang25g.html}.

\bibitem[Kim et~al.(2025{\natexlab{a}})Kim, Finn, and Liang]{kim2025fine}
Moo~Jin Kim, Chelsea Finn, and Percy Liang.
\newblock Fine-tuning vision-language-action models: Optimizing speed and success.
\newblock In \emph{Robotics: Science and Systems (RSS)}, 2025{\natexlab{a}}.
\newblock \doi{10.15607/RSS.2025.XXI.017}.
\newblock URL \url{https://www.roboticsproceedings.org/rss21/p017.html}.

\bibitem[Kim et~al.(2025{\natexlab{b}})Kim, Pertsch, Karamcheti, Xiao, Balakrishna, Nair, Rafailov, Foster, Sanketi, Vuong, Kollar, Burchfiel, Tedrake, Sadigh, Levine, Liang, and Finn]{kim2025openvla}
Moo~Jin Kim, Karl Pertsch, Siddharth Karamcheti, Ted Xiao, Ashwin Balakrishna, Suraj Nair, Rafael Rafailov, Ethan~P. Foster, Pannag~R. Sanketi, Quan Vuong, Thomas Kollar, Benjamin Burchfiel, Russ Tedrake, Dorsa Sadigh, Sergey Levine, Percy Liang, and Chelsea Finn.
\newblock {OpenVLA}: An open-source vision-language-action model.
\newblock In \emph{Proceedings of the 8th Conference on Robot Learning}, volume 270 of \emph{Proceedings of Machine Learning Research}, pages 2679--2713, 2025{\natexlab{b}}.
\newblock URL \url{https://proceedings.mlr.press/v270/kim25c.html}.

\bibitem[Liang et~al.(2023)Liang, Huang, Xia, Xu, Hausman, Ichter, Florence, and Zeng]{liang2023code}
Jacky Liang, Wenlong Huang, Fei Xia, Peng Xu, Karol Hausman, Brian Ichter, Pete Florence, and Andy Zeng.
\newblock Code as policies: Language model programs for embodied control.
\newblock In \emph{2023 IEEE International Conference on Robotics and Automation (ICRA)}, pages 9493--9500, 2023.
\newblock \doi{10.1109/ICRA48891.2023.10160591}.

\bibitem[Liu et~al.(2026)Liu, Li, Yao, Shi, Zhou, Huang, Huang, and Mao]{liu2026guava}
Haowen Liu, Xirui Li, Shaoxiong Yao, Peng Shi, Tianyi Zhou, Jia-Bin Huang, Furong Huang, and Jiayuan Mao.
\newblock Guava: An effective and universal harness for embodied manipulation.
\newblock \emph{arXiv preprint arXiv:2606.18363}, 2026.
\newblock URL \url{https://arxiv.org/abs/2606.18363}.

\bibitem[Lu et~al.(2026)Lu, Wu, Kou, Fu, Xiao, Mandlekar, Xu, Shi, Goldberg, Chen, Chowdhury, Zhu, Fan, and Wang]{aspire2026}
Runyu Lu, Yubo Wu, Ethan Kou, Letian Fu, Wenli Xiao, Ajay Mandlekar, Yinzhen Xu, Guanya Shi, Ken Goldberg, Ang Chen, Mosharaf Chowdhury, Yuke Zhu, Linxi Fan, and Guanzhi Wang.
\newblock {ASPIRE}: Agentic /skills discovery for robotics.
\newblock \emph{arXiv preprint arXiv:2607.00272}, 2026.
\newblock URL \url{https://arxiv.org/abs/2607.00272}.

\bibitem[Mitra et~al.(2025)Mitra, Luo, Saravanan, Niu, Pai, Thomason, Darrell, Anwar, Ramanan, and Herzig]{mitra2025mechanisticfinetuningvisionlanguageactionmodels}
Chancharik Mitra, Yusen Luo, Raj Saravanan, Dantong Niu, Anirudh Pai, Jesse Thomason, Trevor Darrell, Abrar Anwar, Deva Ramanan, and Roei Herzig.
\newblock Mechanistic finetuning of vision-language-action models via few-shot demonstrations.
\newblock \emph{arXiv preprint arXiv:2511.22697}, 2025.
\newblock URL \url{https://arxiv.org/abs/2511.22697}.

\bibitem[{NVIDIA}(2025)]{gr00tn15}
{NVIDIA}.
\newblock {GR00T N1.5}: An improved open foundation model for generalist humanoid robots.
\newblock NVIDIA Research, jun 2025.
\newblock URL \url{https://research.nvidia.com/labs/gear/gr00t-n1_5/}.

\bibitem[{NVIDIA}(2026)]{nvidia2026cosmos3omnimodalworld}
{NVIDIA}.
\newblock Cosmos 3: Omnimodal world models for physical {AI}.
\newblock \emph{arXiv preprint arXiv:2606.02800}, 2026.
\newblock URL \url{https://arxiv.org/abs/2606.02800}.

\bibitem[{OpenAI}(2026)]{openai2026gpt6astra}
{OpenAI}.
\newblock {GPT-6 Astra}: A new generation of intelligence.
\newblock Official model release, 2026.
\newblock URL \url{https://openai.com/index/gpt-6-astra/}.

\bibitem[{Qwen Team}(2026)]{qwen3.8flashnext}
{Qwen Team}.
\newblock {Qwen3.8-Flash-Next}: A new architecture, towards ultimate cost-efficiency, August 2026.
\newblock URL \url{https://qwen.ai/blog?id=qwen3.8-flash-next}.

\bibitem[Shah et~al.(2025)Shah, Chen, Godbole, Mora, Seshia, and Levine]{shah2025liten}
Ameesh Shah, William Chen, Adwait Godbole, Federico Mora, Sanjit~A. Seshia, and Sergey Levine.
\newblock Learning affordances at inference-time for vision-language-action models.
\newblock \emph{arXiv preprint arXiv:2510.19752}, 2025.
\newblock URL \url{https://arxiv.org/abs/2510.19752}.

\bibitem[{Tencent Robotics X} et~al.(2026){Tencent Robotics X}, {HY Vision Team}, Yu, Liu, Wang, Zhang, Rao, Liu, Zhang, Zhao, Wang, Liang, Lin, Wang, Dong, Cheng, Ni, Huang, Hu, Zhang, {Linus}, and Yao]{x2026hyembodied05embodiedfoundationmodels}
{Tencent Robotics X}, {HY Vision Team}, Xumin Yu, Zuyan Liu, Ziyi Wang, He~Zhang, Yongming Rao, Fangfu Liu, Yani Zhang, Ruowen Zhao, Oran Wang, Yves Liang, Haitao Lin, Minghui Wang, Yubo Dong, Kevin Cheng, Bolin Ni, Rui Huang, Han Hu, Zhengyou Zhang, {Linus}, and Shunyu Yao.
\newblock {HY-Embodied-0.5}: Embodied foundation models for real-world agents.
\newblock \emph{arXiv preprint arXiv:2604.07430}, 2026.
\newblock URL \url{https://arxiv.org/abs/2604.07430}.

\bibitem[Wang et~al.(2026)Wang, Yu, Rao, Ling, Li, Wang, Gao, Zhou, Liang, Liu, Zhang, Huang, Xu, Yuan, Yuan, Tan, Zhang, Huang, Zhang, Wu, Hu, and Zhang]{wang2026hyembodiedvlm10efficientphysicalworldagents}
Ziyi Wang, Xumin Yu, Yongming Rao, Yonggen Ling, Yunheng Li, Oran Wang, Mingqi Gao, Yuchen Zhou, Yves Liang, Zuyan Liu, Yani Zhang, Rui Huang, Xiaoran Xu, Bowen Yuan, Yifu Yuan, Xu~Tan, He~Zhang, Yufei Huang, Shenghao Zhang, Hongsheng Wu, Han Hu, and Zhengyou Zhang.
\newblock {HY-Embodied-VLM-1.0}: Efficient physical-world agents.
\newblock \emph{arXiv preprint arXiv:2607.12894}, 2026.
\newblock URL \url{https://arxiv.org/abs/2607.12894}.

\bibitem[{Z.ai}(2026)]{zai2026glm53flash}
{Z.ai}.
\newblock {GLM-5.3-Flash}: More intelligence with less compute.
\newblock Official model release, 2026.
\newblock URL \url{https://autoclaw.z.ai/blog/model/glm-5.3-flash/}.

\bibitem[Zhang et~al.(2026)Zhang, Zhang, Gao, Li, Liu, Nie, Zhu, Qiu, Yan, Liu, Tang, Rao, Fang, Wei, Wang, Ding, and Yu]{zhang2026harnessvla}
Yixian Zhang, Huanming Zhang, Feng Gao, Xiao Li, Zhihao Liu, Yi~Nie, Chunyang Zhu, Jiaxing Qiu, Yuchen Yan, Jiyuan Liu, Wenhao Tang, Jiaji Rao, Zhengru Fang, Changxu Wei, Yu~Wang, Wenbo Ding, and Chao Yu.
\newblock {Harness VLA}: Steering frozen {VLAs} into reliable manipulation primitives via memory-guided agents.
\newblock \emph{arXiv preprint arXiv:2607.08448}, 2026.
\newblock URL \url{https://arxiv.org/abs/2607.08448}.

\bibitem[Zhou et~al.(2025)Zhou, Xu, Tie, Chen, Zhang, Chu, Zhou, and Sun]{zhou2026liberoprorobustfairevaluation}
Xueyang Zhou, Yangming Xu, Guiyao Tie, Yongchao Chen, Guowen Zhang, Duanfeng Chu, Pan Zhou, and Lichao Sun.
\newblock {LIBERO-PRO}: Towards robust and fair evaluation of vision-language-action models beyond memorization.
\newblock \emph{arXiv preprint arXiv:2510.03827}, 2025.
\newblock URL \url{https://arxiv.org/abs/2510.03827}.

\end{thebibliography}

\clearpage
\appendix
\addtocontents{toc}{\protect\setcounter{tocdepth}{3}}
\phantomsection
\hypertarget{toc}{}
\tableofcontents
\clearpage

\section{Related Work}
\label{app:related-work}

\subsection{Robot Foundation Models}
Vision-language-action (VLA) models adapt pretrained vision-language models
into robot policies and map observations and language instructions directly to low-level actions~\citep{brohan2023rt2,kim2025openvla,octomodelteam2024octoopensourcegeneralistrobot,black2024pi0visionlanguageactionflowmodel}.
More recent systems improve continuous action generation and open-world transfer by leveraging the rapid growth of robot datasets and
advances in generative modeling~\citep{black2025pi05,gr00tn15,molmoact2_2026}.

Despite their broader task coverage, the performance of robot foundation policies
remains bounded by the coverage of their robot-action training distribution.
Deployment to new task settings and embodiments commonly requires collecting
in-domain demonstrations and fine-tuning the policy,
limiting truly zero-shot transfer\citep{kim2025fine, mitra2025mechanisticfinetuningvisionlanguageactionmodels}.
Moreover, adapting a VLM to low-level actions can weaken the
general multimodal understanding, instruction following, and semantic reasoning capabilities
that originally motivates the VLA paradigm \citet{hancock2025actionslanguagefinetuningvlms}.

In contrast, we explore zero-shot robot control with a general-purpose VLM, without task-specific robot data or policy fine-tuning, and directly leverage its spatial priors and semantic knowledge through an explicit primitive action interface.

\subsection{Coding Agents for Robot Manipulation}
Language models can also control robots by generating programs over robot APIs.
Prior works use language models to generate executable robot programs or
spatial objectives and constraints that are passed to motion
planners~\citep{liang2023code,huang2023voxposer,huang2024rekep}.
More recent work extends this idea in different ways. CaP-X studies agents that generate, execute, and revise
robot programs, ASPIRE acquires reusable skills through
exploration, and RHO uses execution feedback to optimize multi-file control
software before deployment~\citep{fu2026capx,aspire2026,elmaaroufi2026rho}.

However, coding-agent systems face a trade-off between responsiveness and efficiency. Online code revision introduces the latency of multi-turn program generation, while offline variants rely on programs or skills fixed before deployment and cannot adapt to unexpected environmental changes during execution.

\subsection{VLM-Based Robot Control}
Recent systems combine a high-level VLM with a learned robot policy. LITEN
reflects on past execution trajectories to learn which instructions a low-level
VLA can perform and uses this experience to improve later plans~\citep{shah2025liten}.
VoLo and Harness VLA similarly use VLMs for planning and orchestration while relying on frozen VLA
policies and analytic primitives for physical execution~\citep{chen2026volo,zhang2026harnessvla}.
These systems add reasoning around learned robot policies, but their physical capabilities still depend on the underlying action models and therefore can't be readily deployed in new settings without adapting the action backend.

Another line of work enables general-purpose VLMs to control robots through compact interfaces composed of translation, rotation, gripper, and other semantic actions~\citep{hu2026via,guo2026robodawn,liu2026guava}.~\citep{hu2026via,guo2026robodawn,liu2026guava}.

Concurrent with our work, Show-Harness introduces a modular, plugin-based framework that augments a sequential observe--reason--act loop with situated planning, action chunking, and predefined failure-recovery routines~\citep{chen2026showharness}. Both works are motivated by a similar intuition: enabling general-purpose VLMs to control robots through an interface resembling human teleoperation. However, the resulting harness designs differ substantially. First, execution scheduling differs: Show-Harness follows a primarily sequential interaction loop, whereas \MethodName{} runs monitoring concurrently with physical execution and generates memory summaries in the background, while action proposals, outcome assessment, and replanning remain sequential. Second, decision autonomy differs: Show-Harness formulates control largely as VLM-based tool selection over predefined Python plugins, while \MethodName{} explicitly decomposes decision making across planning, execution, monitoring, and verification, allowing the VLM to reason directly about task decomposition and intermediate decisions. Third, motion parameterization differs: Show-Harness grounds actions through predefined or rule-selected motion increments, whereas our Executor predicts parameterized mid-level actions with VLM-reasoned motion magnitudes. Finally, correction mechanisms differ: Show-Harness primarily performs reactive recovery after execution failures, while \MethodName{} supports proactive interruption by monitoring ongoing execution and canceling pending actions when they become inconsistent with the evolving scene. Because Show-Harness appeared concurrently with our work, we focus on clarifying these methodological distinctions rather than providing a direct experimental comparison.

\section{Diagnostic Evaluation Details}
\label{app:diagnostic}

\subsection{Data Construction}
\label{app:diagnostic-data}

The diagnostic seperates three local decisions that arise during manipulation:
selecting what to do next, deciding whether an observed transition advances the
current subgoal, and deciding whether the current state satisfies that subgoal.
Separating these questions prevents a single episode-level success indicator
from obscuring their different failure modes. The unit of evaluation is
therefore one image-conditioned question about a local subgoal, rather than a
closed-loop execution of a complete task. Table~\ref{tab:diagnostic-task-summary}
summarizes the resulting 240-question set.

\begin{table}[t]
    \centering
    \small
    \setlength{\tabcolsep}{4pt}
    \renewcommand{\arraystretch}{1.08}
    \begin{tabularx}{\linewidth}{lXlc}
        \toprule
        \textbf{Task} & \textbf{Decision being diagnosed} & \textbf{Visual input} & \textbf{Count} \\
        \midrule
        Action selection
        & Primary primitive to execute after the current observation $t_1$
        & Scene and wrist views at $t_0,t_1$ & 80 \\
        Progress verification
        & Whether the transition from $t_0$ to $t_1$ advances the stated subgoal
        & Scene and wrist views at $t_0,t_1$ & 80 \\
        Subsubgoal completion
        & Whether the state at $t_1$ satisfies the stated subgoal and criterion
        & Scene and wrist views at $t_1$ & 80 \\
        \bottomrule
    \end{tabularx}
    \caption{\textbf{Composition of the manipulation decision diagnostic.}}
    \label{tab:diagnostic-task-summary}
\end{table}

We constructed the questions from LIBERO expert demonstrations. Codex inspected
the expert trajectories and their private state evidence, selected temporal
windows, and decomposed each task into local decisions. Rather than sampling
frames uniformly, it selected segments with a visually discernible change and
a meaningful decision at $t_1$, then adjusted $t_0$ and $t_1$ when the change
was too small or the interval extended beyond the relevant decision.

Codex also rewrote complete task instructions into local, observable subgoals
and success criteria, such as approaching a handle, aligning the gripper,
grasping or lifting an object, carrying it above a destination, lowering and
releasing it, opening or closing a drawer, and changing an object's
orientation. A single visual segment may support both an action question and a
progress question, whereas completion questions use states selected for judging
the result. Exact text deduplication gives 138 distinct subgoal strings
and 142 distinct \texttt{(subgoal, success criterion)} pairs.

\subsection{Question Generation and Labels}
\label{app:diagnostic-questions}

Question construction jointly specifies the observation window, a local
subgoal, its observable success criterion, and the reference answer. During
annotation, Codex had access not only to the two camera views shown in each
question, but also to private construction-time evidence: end-effector and
object poses, joint and gripper states, expert controls, short future windows,
simulator predicates, and the outcomes of perturbed and control rollouts.

Codex proposed the initial questions, labels, and refines by jointly
inspecting the images and private evidence. These proposals were then reviewed
by a human expert, whose feedback was used to correct labels, replace ambiguous frames,
revise object relations, and narrow or expand subgoal scope.

The action task asks for one primary next action and uses three JSON schemas:
\texttt{move} with \texttt{forward}, \texttt{backward}, \texttt{left},
\texttt{right}, \texttt{up}, or \texttt{down}; \texttt{rotate} with
\texttt{yaw\_left}, \texttt{yaw\_right}, \texttt{pitch\_up},
\texttt{pitch\_down}, \texttt{roll\_ccw}, or \texttt{roll\_cw}; and
\texttt{gripper} with state \texttt{open} or \texttt{close}. Both translations and rotations are
expressed in the robot base frame using the right-hand rule. The label specifies
only the prioritized primitive, not a distance, angle, speed, or complete
executable command.

We determined each reference action from the local subgoal together with the
next expert control, the end-effector displacement over a short window, and
changes in object and grasp state. For rotational evidence, we computed
relative orientation as
\begin{equation}
    \Delta R = R_1 R_0^{-1},
\end{equation}
converted it to a rotation vector, and expressed that vector in the robot base
frame.

Progress is labeled as a function of
$(O_{t_0},O_{t_1},\text{subgoal},\text{criterion})$. A transition receives
label 1 when it advances the specified subgoal and label 0 when it does not,
acts on the wrong object, disrupts a required relation, or moves toward an
incompatible state.We construct negative progress examples either by pairing an expert transition
with a visually grounded counterfactual subgoal, or by restoring an expert
state and executing a physically simulated action perturbation. We retain only
perturbations whose realized outcomes do not advance the specified subgoal.

Subsubgoal completion is judged only from the two views at $t_1$ and the explicitly
stated success criterion. Label 1 means that the criterion is satisfied; label
0 means that it remains incomplete or that the scene is in an incompatible
state. Completion negatives capture two distinct cases: a goal or state
mismatch, or correct progress that has not yet met the completion criterion.

\subsection{Prompts and Scoring}
\label{app:diagnostic-prompts}

We use the following prompts for the three diagnostic tasks.

\newtcblisting{diagnosticprompt}[1]{
    enhanced,
    breakable,
    colback=gray!4!white,
    colframe=black,
    boxrule=0.4pt,
    arc=0pt,
    outer arc=0pt,
    left=2mm,
    right=2mm,
    top=1.5mm,
    bottom=1.5mm,
    listing only,
    listing options={
        basicstyle=\ttfamily\small,
        breaklines=true,
        columns=fullflexible,
        keepspaces=true,
        showstringspaces=false
    },
    before={\par\medskip\noindent\textbf{#1 Prompt}\par\smallskip},
    after={\par\medskip}
}

\begin{diagnosticprompt}{Action Selection}
You are evaluating robot observations for an embodied decision task. Use only
the supplied images, subgoal and success criterion. Return exactly one JSON
object with no explanation.
Images are upright RGB. The images labeled "scene" are from the fixed EXTERNAL
camera; the images labeled "wrist" are from the camera mounted on the gripper,
which moves with it. Camera motion alone is not object motion. Compare the two
views and, when supplied, the two times.
SUBGOAL REFERENCE FRAME: Unless explicitly stated otherwise, left/right and
leftmost/rightmost in the natural-language subgoal or success criterion refer
to LEFT and RIGHT IN THE FIXED EXTERNAL CAMERA IMAGE (the view labeled
"scene"), as seen by the viewer looking at the displayed image. Image-left
means the left side of that image; image-right means the right side of that
image. For example, "the bowl to the right of the plate" identifies the bowl
on the plate's image-right side in the scene view. These words do not specify a
robot-base action direction. Do not use the moving wrist camera to redefine
these relations. Object-relative descriptions such as "in front of the stove"
retain their contextual spatial meaning; do not automatically reinterpret them
as base +X or -X.
ACTION OUTPUT REFERENCE FRAME: Only the direction names in the output action
use the robot BASE coordinate system. Translation: forward=+X, backward=-X,
left=+Y, right=-Y, up=+Z, down=-Z. Rotations use the right-hand rule about base
axes: yaw_left=+Z, yaw_right=-Z, pitch_up=-Y, pitch_down=+Y, roll_ccw=+X,
roll_cw=-X. Base and world axes align in these scenes. A target on image-right
does not necessarily require the action "right"; identify the intended target
in the scene view, then choose the base-frame motion toward it. The reference
frames are distinct; no universal image-left/base-left or
image-right/base-right equivalence is specified.
For action selection, choose the primary NEXT primitive after current t1 that
advances the subgoal; do not describe the past motion. Valid outputs:
{"type":"move","direction":"forward|backward|left|right|up|down"},
{"type":"rotate","direction":"yaw_left|yaw_right|pitch_up|pitch_down|roll_ccw|roll_cw"},
or {"type":"gripper","state":"open|close"}. Choose a
single value, not the literal pipe-separated list. No distance or angle is
requested.
\end{diagnosticprompt}

\begin{diagnosticprompt}{Progress Verification}
You are evaluating robot observations for an embodied decision task. Use only
the supplied images, subgoal and success criterion. Return exactly one JSON
object with no explanation.
Images are upright RGB. The images labeled "scene" are from the fixed EXTERNAL
camera; the images labeled "wrist" are from the camera mounted on the gripper,
which moves with it. Camera motion alone is not object motion. Compare the two
views and, when supplied, the two times.
SUBGOAL REFERENCE FRAME: Unless explicitly stated otherwise, left/right and
leftmost/rightmost in the natural-language subgoal or success criterion refer
to LEFT and RIGHT IN THE FIXED EXTERNAL CAMERA IMAGE (the view labeled
"scene"), as seen by the viewer looking at the displayed image. Image-left
means the left side of that image; image-right means the right side of that
image. For example, "the bowl to the right of the plate" identifies the bowl
on the plate's image-right side in the scene view. These words do not specify a
robot-base action direction. Do not use the moving wrist camera to redefine
these relations. Object-relative descriptions such as "in front of the stove"
retain their contextual spatial meaning; do not automatically reinterpret them
as base +X or -X.
For progress verification, output {"label":1} if the change from t0 to t1
advances the stated subgoal, otherwise {"label":0}. A subgoal need not be
complete to show progress; preparation counts when it advances the stated
subgoal. Mere unrelated motion does not.
\end{diagnosticprompt}

\begin{diagnosticprompt}{Subgoal Completion}
You are evaluating robot observations for an embodied decision task. Use only
the supplied images, subgoal and success criterion. Return exactly one JSON
object with no explanation.
Images are upright RGB. The images labeled "scene" are from the fixed EXTERNAL
camera; the images labeled "wrist" are from the camera mounted on the gripper,
which moves with it. Camera motion alone is not object motion. Compare the two
views and, when supplied, the two times.
SUBGOAL REFERENCE FRAME: Unless explicitly stated otherwise, left/right and
leftmost/rightmost in the natural-language subgoal or success criterion refer
to LEFT and RIGHT IN THE FIXED EXTERNAL CAMERA IMAGE (the view labeled
"scene"), as seen by the viewer looking at the displayed image. Image-left
means the left side of that image; image-right means the right side of that
image. For example, "the bowl to the right of the plate" identifies the bowl
on the plate's image-right side in the scene view. These words do not specify a
robot-base action direction. Do not use the moving wrist camera to redefine
these relations. Object-relative descriptions such as "in front of the stove"
retain their contextual spatial meaning; do not automatically reinterpret them
as base +X or -X.
For subgoal completion, output {"label":1} only if the stated subgoal and its
success criterion are satisfied at t1, otherwise {"label":0}. If release is
required, holding an object at the target is insufficient.
\end{diagnosticprompt}

Table~\ref{tab:action-type-breakdown} reports action-selection accuracy
separately for translation, rotation, and gripper primitives.

\begin{table*}[t]
    \centering
    \scriptsize
    \setlength{\tabcolsep}{6.5pt}
    \renewcommand{\arraystretch}{1.15}
    \begin{tabular}{lccc}
        \toprule
        Model & Translation (48) & Rotation (19) & Gripper (13) \\
        \midrule
        Qwen3.8-Flash-Next-FP8 & 47.92\% (23/48) & 0.00\% (0/19) & 46.15\% (6/13) \\
        HY-Embodied-0.5 MoT-2B & 20.83\% (10/48) & 15.79\% (3/19) & 23.08\% (3/13) \\
        Hy-Embodied-VLM-1.0 A3B & 31.25\% (15/48) & 0.00\% (0/19) & 0.00\% (0/13) \\
        Cosmos3-Nano (understanding tower) & 27.08\% (13/48) & 0.00\% (0/19) & 15.38\% (2/13) \\
        GLM-5.3-Flash & 41.67\% (20/48) & 15.79\% (3/19) & 53.85\% (7/13) \\
        GPT-6 Astra (medium) & 58.33\% (28/48) & 63.16\% (12/19) & 61.54\% (8/13) \\
        \bottomrule
    \end{tabular}
    \caption{\textbf{Action-selection results by primitive type.} Accuracy is
    reported separately for translation, rotation, and gripper questions.}
    \label{tab:action-type-breakdown}
\end{table*}

\section{MotorMind Implementation Details}
\label{app:motormind-implementation}

\subsection{VLM Roles and Prompts}
\label{app:motormind-vlm-roles-prompts}

MotorMind separates task-level reasoning, action generation, execution
monitoring, outcome verification, and memory construction through role-specific
VLM calls. The separation is enforced by both the information supplied to each
call and its structured output schema. In particular, the executor receives one
subgoal rather than the full planning problem, while the motion supervisor and
scene monitor have no action field and therefore cannot silently replace an
action already being executed.

\newtcolorbox{vlmroleprompt}[1]{
    enhanced,
    breakable,
    colback=gray!4!white,
    colframe=black,
    boxrule=0.4pt,
    arc=0mm,
    outer arc=0mm,
    fontupper=\ttfamily\small,
    left=2mm,
    right=2mm,
    top=1.5mm,
    bottom=1.5mm,
    before={\par\medskip\noindent\textbf{#1 Prompt}\par\smallskip},
    after={\par\medskip}
}

\begin{vlmroleprompt}{Planner}
You turn a task told in one sentence into a list of small subgoals for a robot
arm. You do not move the robot: another agent, the executor, takes one subgoal
at a time and proposes the motions itself. Everything you have worked out must
reach it as the wording of a subgoal, because the wording and the criterion are
the only things it is given. You answer with one JSON object and nothing else.
\end{vlmroleprompt}

\begin{vlmroleprompt}{Replanner}
A plan you wrote is part-way through and something has gone wrong. You write the
steps that are still to come. The finished ones are kept by the caller and must
not appear again. You answer with one JSON object and nothing else.
\end{vlmroleprompt}

\begin{vlmroleprompt}{Executor}
You decide what a robot arm should do next. You are given a subgoal, what the
cameras see now, and what the robot reports about itself. You answer with one
JSON object and nothing else. You are not the planner: choose the next few
actions that move the subgoal forward, not the whole task. They may be a move
across the table, a descent, a lift, a turn, a gripper command or a wait --
whatever the situation in front of you actually calls for.
\end{vlmroleprompt}

\begin{vlmroleprompt}{Motion Supervisor}
You are watching one motion that is already running on a robot arm. Your only
decision is whether it should carry on or stop now. You cannot start, change,
extend or replace a motion. You answer with one JSON object and nothing else.
\end{vlmroleprompt}

\begin{vlmroleprompt}{Scene Monitor}
You are watching a robot arm while one step of a task runs. You are the only one
looking at the task as a whole; another agent is already judging each individual
motion, so that is not your job. You cannot start, stop, change or suggest a
motion. You answer with one JSON object and nothing else.
\end{vlmroleprompt}

\begin{vlmroleprompt}{Verifier}
You decide whether one step of a robot task actually happened. You are shown the
scene before the step and the scene now, and what the robot that performed it
claims. You are not that robot and you are not obliged to agree with it. You
answer with one JSON object and nothing else.
\end{vlmroleprompt}

\begin{vlmroleprompt}{Memory Manager}
You keep the memory of a robot run. You are given what the robot was asked to do,
the plan it is following, the note you wrote last time, and the log of what has
happened since -- one line per motion, with the numbers the robot measured. You
answer with one JSON object and nothing else. You do not decide what the robot
does next and you never write an action: you write down what is now known, so
that the planner can decide better.
\end{vlmroleprompt}

\begin{vlmroleprompt}{Target Locator}
You look at pictures of a table and say what is on it and where. You answer with
one JSON object and nothing else. You never describe a motion and you never
mention the robot.
\end{vlmroleprompt}

The prompts above are incomplete excerpts that summarize the role definitions.
We will provide the complete prompt files in the supplementary material.

\subsection{Visual Grounding and Measured Feedback}
\label{app:motormind-grounding}

The Executor receives current images together with a structured description of robot state and recent execution. The state includes tool pose, gripper opening and holding state, measured motion since the previous observation, and available clearance information. Its local history records recent proposals, refusals, and outcomes. When localization is required, the VLM identifies the target through labeled image-space bounding boxes. Compatible fixed-camera views are triangulated. A close wrist view can refine the estimate. The resulting target offsets, uncertainty, and view disagreement are supplied to the Executor in the same base-frame direction vocabulary used by its actions.

\subsection{Structured Proposals and Auxiliary Actions}
\label{app:motormind-actions}

An Executor proposal contains \texttt{assessment}, \texttt{done}, an optional \texttt{command}, \texttt{expect}, and \texttt{confidence}. The command contains an ordered \texttt{actions} list. In addition to the core primitives in Section~\ref{sec:mid-level-actions}, the interface supports \texttt{wait}, with a positive duration, and \texttt{home}. A translation supplies exactly one direction word or base-frame axis and a positive distance in millimeters; a rotation supplies a direction or axis and an angle in degrees. Optional command defaults specify translational and rotational speed, contact behavior, and a note. Gripper actions specify \texttt{open} or \texttt{close}, with an optional width. A completion signal without a command ends the attempt for assessment; with a command, it requests execution of that command before the attempt ends. Schema and conversion errors are returned to the proposer for correction before execution.

\section{Simulation Evaluation Details}
\label{app:simulation-evaluation}

\subsection{LIBERO-Pro Tasks and Perturbations}
\label{app:libero-pro-tasks}

We evaluate three LIBERO task families: Spatial, Object, and Goal.
Each family contains 10 tasks evaluated under five conditions: Base, Language,
Object, Position Swap, and Task. Base is the unperturbed reference condition;
the other four conditions apply the corresponding benchmark-provided
perturbations.

Language perturbations vary the wording of the task instruction. Object
perturbations modify scene objects according to the benchmark definitions.
Position Swap perturbations alter object placements, while Task perturbations
change the instructed target or goal. We use the provided configurations for
each condition. Table~\ref{tab:libero-pro-conditions} summarizes the evaluation
conditions.

The full evaluation matrix comprises $4$ task families $\times$ $5$ conditions
$\times$ $10$ tasks, yielding 200 task configurations: 40 base configurations
and 160 perturbed configurations. These are task configurations rather than
rollout counts; evaluation across multiple initialization seeds produces
additional rollouts. Success is determined by the environment's predicate-based
checks for the evaluated task.

\begin{table}[t]
    \centering
    \small
    \setlength{\tabcolsep}{5pt}
    \renewcommand{\arraystretch}{1.1}
    \begin{tabularx}{\linewidth}{l l X}
        \toprule
        \textbf{Condition} & \textbf{Configuration key} & \textbf{Description} \\
        \midrule
        Base          & \texttt{base}   & Original, unperturbed task setting \\
        Language      & \texttt{lan}    & Instruction wording perturbations \\
        Object        & \texttt{object} & Benchmark-defined object perturbations \\
        Position Swap & \texttt{swap}   & Object placement perturbations \\
        Task          & \texttt{task}   & Target or goal perturbations \\
        \bottomrule
    \end{tabularx}
    \caption{\textbf{LIBERO-Pro evaluation conditions.} Each condition is
    applied to every evaluated task family.}
    \label{tab:libero-pro-conditions}
\end{table}

\subsection{Baseline Configurations}
\label{app:simulation-baselines}

We evaluate direct vision-language-action (VLA) policies and agent frameworks
that combine planning with learned or programmatic execution. The direct VLA
baselines comprise OpenPI $\pi_{0.5}$\citep{black2025pi05}, OpenVLA\citep{kim2025openvla}, OpenVLA-OFT\citep{kim2025fine}, GR00T N1.5\citep{gr00tn15}, and
MolmoAct2\citep{molmoact2_2026}. The agent baselines comprise Harness VLA\citep{zhang2026harnessvla}, VoLo\citep{chen2026volo}, and CAP-X\citep{fu2026capx}. All
evaluated VLA weights remain frozen; we perform no additional policy training
or fine-tuning. For all LIBERO-Pro configurations that use
$\pi_{0.5}$, including its use within Harness VLA and VoLo, we use the same
OpenPI \texttt{libero\_base} checkpoint. Table~\ref{tab:simulation-baselines}
summarizes their control and planning mechanisms.

\begin{table*}[t]
    \centering
    \small
    \setlength{\tabcolsep}{5pt}
    \renewcommand{\arraystretch}{1.1}
    \begin{tabularx}{\textwidth}{l X X}
        \toprule
        \textbf{Baseline} & \textbf{Control mechanism} & \textbf{Planning and memory} \\
        \midrule
        OpenPI $\pi_{0.5}$ & Direct VLA action prediction & No external planner or task memory \\
        OpenVLA & Autoregressive action-token prediction & No external planner or task memory \\
        OpenVLA-OFT & Continuous action-chunk prediction & No external planner or task memory \\
        GR00T N1.5 & Native action-chunk prediction & No external planner or task memory \\
        MolmoAct2 & Continuous action prediction & No external planner or task memory \\
        Harness VLA & Tool-based execution with frozen $\pi_{0.5}$ & VLM planning and task-specific memory \\
        VoLo & VLM-guided execution with frozen $\pi_{0.5}$ & VLM planning with visual perception tools \\
        CAP-X & Generated Python programs over robot tools & Program generation and visual execution feedback \\
        \bottomrule
    \end{tabularx}
    \caption{\textbf{Baseline configurations.} Direct policies predict robot
    actions, whereas agent frameworks add planning, tools, or programmatic
    execution around their control policy.}
    \label{tab:simulation-baselines}
\end{table*}

\paragraph{Direct VLA Policies: Shared Setup.}
Direct VLA policies receive the task instruction and each model's supported
observations, without MotorMind-generated subgoals, memory, or action
corrections. We preserve model-specific image preprocessing, state encoding,
action decoding, and normalization. Predicted actions are converted to LIBERO's
seven-dimensional control interface. Each policy follows its configured action
execution schedule before acquiring a new observation; chunk lengths are not
assumed to be identical across architectures.

\paragraph{OpenPI $\pi_{0.5}$.}
We use the OpenPI \texttt{libero\_base} $\pi_{0.5}$ checkpoint as the direct
$\pi_{0.5}$ baseline and as the shared $\pi_{0.5}$ execution policy in the agent
frameworks. The policy receives visual observations, the language instruction,
and robot state, and predicts continuous action chunks. We retain the
corresponding LIBERO observation and action transforms and normalization. In
direct evaluation, the policy is conditioned on the task instruction; within an
agent framework, it receives the instruction selected by that framework. Its
weights remain frozen in both cases.

\paragraph{OpenVLA and OpenVLA-OFT.}
OpenVLA uses the published \texttt{openvla/openvla-7b} checkpoint with its
autoregressive action decoder. It receives a single scene image and the task
instruction, without proprioceptive input, and executes one action before the
next observation. The implementation retains the original action decoder and
uses suite-specific LIBERO normalization statistics.

OpenVLA-OFT uses the published
\texttt{moojink/openvla-7b-oft-finetuned-libero-}\linebreak
\texttt{spatial-object-goal-10}
checkpoint with its continuous L1 action head and proprioceptive projector. It
receives scene and wrist images, the instruction, and robot state, and executes
eight-action chunks. We retain the model-specific image preprocessing,
proprioceptive normalization, and LIBERO action transforms. Both policies are
evaluated without additional weight updates.

\paragraph{GR00T N1.5.}
For the LIBERO-adapted configuration, we use
\path{youliangtan/gr00t-n1.5-libero-spatial-posttrain} across task families,
without switching to a different checkpoint for each family. The policy
receives scene and wrist images together with robot state and language. We
retain its native LIBERO normalization and eight denoising steps. It predicts
16 actions, of which the first is executed before observing again.

The raw-base configuration uses \texttt{nvidia/GR00T-N1.5-3B} with its
pretrained DROID embodiment and statistics. A DROID-to-LIBERO adapter converts
the observation and action interfaces, and 16 actions are executed between
observations. This configuration evaluates embodiment transfer and is distinct
from the LIBERO-adapted policy; its results retain a separate baseline label.

\paragraph{MolmoAct2.}
We use the published \texttt{allenai/MolmoAct2} and
\texttt{allenai/MolmoAct2-LIBERO} checkpoints for the base and LIBERO-adapted
configurations, respectively. Both use continuous-action inference with ten
flow steps, normalized language, and no depth-reasoning mode. The input contains
scene and wrist images, the task instruction, and robot state; ten predicted
actions are executed before the next observation.

The LIBERO-adapted policy uses its own LIBERO statistics. The base-weight
configuration uses LIBERO-derived normalization and robot metadata while
retaining the base model weights. Accordingly, the base configuration involves
no additional policy training, but is not free of LIBERO-derived interface
information. These VLA policy can also be the ablation for mid-level action representation.

\paragraph{Harness VLA.}
Harness VLA combines a Qwen3.8-Flash-Next-FP8 planner with perception and
execution tools, including SAM3 and the frozen OpenPI \texttt{libero\_base}
$\pi_{0.5}$ policy. The planner selects tool calls and execution instructions,
while the VLA provides the learned manipulation component. We retain the
framework's native tool interface rather than replacing it with MotorMind's
semantic-action executor.

Harness VLA additionally constructs task-specific memory through exploration.
Each task starts from empty memory on seed 0, and the resulting memory is frozen
at the designated exploration budget before evaluation on new initial states.
The evaluated memory-budget configurations use snapshots after five or ten
exploration attempts. Exploration changes the stored memory rather than the VLA
weights. Exploration and evaluation costs are treated separately, and memory
budgets are identified with their corresponding results.

\paragraph{VoLo.}
VoLo retains its Qwen3.8-Flash-Next-FP8 planner and SAM3/Molmo perception tools.
The planner uses visual context and tool outputs to guide execution by the
frozen OpenPI \texttt{libero\_base} $\pi_{0.5}$ policy. We retain the
framework's LIBERO observation and action transforms and normalization, with
five actions executed before the next policy observation. No additional policy
training is performed. VoLo is evaluated as an agent framework with its own
planning and perception workflow; it does not receive MotorMind's memory,
task-clause ledger, or geometric action corrections.
VoLo is sequential execution agent, which can also serve as the ablation for our asynchronous harness system.

\paragraph{CAP-X.}
CAP-X is a program-generating baseline. It retains the agent0 \texttt{ReducedSkillLibrary} interface and generates executable Python code
from visual observations and the task instruction. Generated programs invoke
the available robot tools, and subsequent visual feedback supports correction.
Unlike the VLA-based frameworks above, CAP-X does not use $\pi_{0.5}$ as its
execution policy.

Our implementation provides RGB-D observations and uses joint-position
control. Candidate program generation uses GPT-5.6-terra and Claude-Opus-5,
with four candidates from each model; GPT-5.6-terra performs synthesis and
visual differencing. The method retains its code-generation and feedback
workflow rather than adopting MotorMind's planner or action interface.
Evaluation uses the applicable environment task definition and predicate-based
success checks.

\section{Adaptive Reasoning Tasks}
\label{app:adaptive-reasoning}

We evaluate adaptive reasoning in four online manipulation suites comprising 30
tasks. The suites test responses to changes in object location, continuously
moving targets, semantic and temporal target-selection requirements, and
revised instructions. Task definitions and intervention logic are implemented
in the environment. Object identities and state variables used to trigger
interventions or score outcomes are not supplied as privileged policy inputs.
This study be framed as stress tests rather than precise quantitative evidence.

\subsection{Details of the Four Task Suites}
\label{app:adaptive-task-suites}

Table~\ref{tab:adaptive-suite-summary} summarizes the four suites and their
evaluation focus.

\begin{table}[t]
    \centering
    \small
    \setlength{\tabcolsep}{5pt}
    \renewcommand{\arraystretch}{1.1}
    \begin{tabularx}{\linewidth}{l c X}
        \toprule
        \textbf{Suite} & \textbf{Tasks} & \textbf{Evaluation focus} \\
        \midrule
        Interactive & 10 & Adapt to object or receiver displacement with an unchanged instruction. \\
        Pure Dynamic & 5 & Identify, intercept, and place a named moving target. \\
        Dynamic Reasoning & 10 & Resolve semantic, relational, and temporal references to moving targets. \\
        Prompt Change & 5 & Respond to revised instructions without resetting the physical scene. \\
        \bottomrule
    \end{tabularx}
    \caption{\textbf{Adaptive reasoning task suites.} The four suites contain
    30 online manipulation tasks in total.}
    \label{tab:adaptive-suite-summary}
\end{table}

\paragraph{Scene Shift.}
The instruction remains fixed while the environment displaces either the object
to be grasped or its receiving container. Interventions are triggered by
physical progress, such as approaching an object before grasping or approaching
a receiver after lifting the payload. Tasks include one, two, or three
displacements, testing both initial adaptation and repeated correction. The
number of displacements in Table~\ref{tab:interactive-tasks} specifies the
configured maximum; later interventions require the corresponding progress
triggers to be reached.

\begin{table*}[t]
    \centering
    \scriptsize
    \setlength{\tabcolsep}{4pt}
    \renewcommand{\arraystretch}{1.08}
    \begin{tabularx}{\textwidth}{c X X}
        \toprule
        \textbf{Task} & \textbf{Instruction} & \textbf{Intervention} \\
        \midrule
        1 & put the ketchup in the wooden tray & Receiver during transport; 1 displacement \\
        2 & put the bowl on the plate & Receiver during transport; 1 displacement \\
        3 & put the wine bottle on the rack & Target object before grasping; 1 displacement \\
        4 & put the cream cheese in the bowl & Receiver during transport; 1 displacement \\
        5 & put the bbq sauce in the wooden tray & Target object before grasping; 2 displacements \\
        6 & put the milk in the basket & Receiver during transport; 2 displacements \\
        7 & put the chocolate pudding in the bowl & Receiver during transport; 2 displacements \\
        8 & put the salad dressing in the wooden tray & Target object before grasping; 3 displacements \\
        9 & put the orange juice in the basket & Target object before grasping; 3 displacements \\
        10 & put the butter in the bowl & Receiver during transport; 3 displacements \\
        \bottomrule
    \end{tabularx}
    \caption{\textbf{Scene shift tasks.} The instruction remains unchanged
    while the target object or receiving container is displaced.}
    \label{tab:interactive-tasks}
\end{table*}

\paragraph{Dynamic Manipulation.}
These tasks use a single-pass conveyor. The instruction names the target, and
the agent must locate it among the moving objects, grasp it, and place it in the
designated receiver. The selected configurations include conveyor motion at
1.5\,mm/s. Objects follow scripted conveyor motion until a two-finger grasp
transfers them to ordinary rigid-body dynamics. Targets that are not grasped can
leave the available conveyor region; the motion does not loop to provide
repeated passes.

\begin{table*}[t]
    \centering
    \scriptsize
    \setlength{\tabcolsep}{4pt}
    \renewcommand{\arraystretch}{1.08}
    \begin{tabularx}{\textwidth}{c X}
        \toprule
        \textbf{Task} & \textbf{Instruction} \\
        \midrule
        1 & pick up the red mug from the conveyor belt and place it in the basket \\
        2 & pick up the alphabet soup from the conveyor belt and place it in the basket \\
        3 & pick up the ketchup from the conveyor belt and place it in the red basket \\
        4 & pick up the blue-and-white cream cheese box from the conveyor belt and place it in the wooden tray \\
        5 & pick up the green ketchup from the conveyor belt and place it in the basket \\
        \bottomrule
    \end{tabularx}
    \caption{\textbf{Dynamic manipulation tasks.} Each instruction identifies a named
    target moving on a single-pass conveyor.}
    \label{tab:pure-dynamic-tasks}
\end{table*}

\paragraph{Dynamic Reasoning.}
These tasks combine conveyor manipulation with target selection from semantic
properties, exclusion, spatial relations, or observation history. Spatial
references explicitly refer to the initial arrangement.

\begin{table*}[t]
    \centering
    \scriptsize
    \setlength{\tabcolsep}{4pt}
    \renewcommand{\arraystretch}{1.08}
    \begin{tabularx}{\textwidth}{c l X}
        \toprule
        \textbf{Task} & \textbf{Reasoning requirement} & \textbf{Instruction} \\
        \midrule
        1 & Contents: liquid & pick up the packaged product containing liquid from the conveyor belt and place it in the basket \\
        2 & Category: edible item & pick up the edible item from the conveyor belt and place it in the basket \\
        3 & Purpose: beverage & pick up the beverage meant for drinking from the conveyor belt and place it in the basket \\
        4 & Category exclusion: not a bowl & pick up the drinking vessel that is not a bowl from the conveyor belt and place it in the basket \\
        5 & Attribute exclusion: not red & pick up the cup that is not red from the conveyor belt and place it in the basket \\
        6 & Initial spatial relation & pick up the object that is between the two bowls in the initial arrangement on the conveyor belt and place it in the basket \\
        7 & Cumulative temporal counting & Select the second distinct food item over the entire episode, then pick it up downstream of the yellow line and place it in the basket. \\
        8 & Purpose: seasoning, not drinking & pick up the bottled product used to season food rather than to drink from the conveyor belt and place it in the basket \\
        9 & Contents: fruit drink & pick up the drink made from fruit from the conveyor belt and place it in the basket \\
        10 & Category and initial spatial relation & pick up the food item that is between the two cups in the initial arrangement on the conveyor belt and place it in the basket \\
        \bottomrule
    \end{tabularx}
    \caption{\textbf{Dynamic reasoning tasks.} The target is specified through
    semantic, exclusion-based, spatial, or temporal constraints.}
    \label{tab:dynamic-reasoning-tasks}
\end{table*}

\paragraph{Prompt Shift.}
The environment replaces the instruction after a predefined physical trigger
while preserving the robot and scene state. The tasks cover destination
redirection, insertion of an intermediate operation, an obstacle reminder, and
target-object replacement. The agent must adapt from its current physical state
rather than restart the original task.

\begin{table*}[t]
    \centering
    \scriptsize
    \setlength{\tabcolsep}{4pt}
    \renewcommand{\arraystretch}{1.08}
    \begin{tabularx}{\textwidth}{c X X}
        \toprule
        \textbf{Task} & \textbf{Initial instruction} & \textbf{Replacement instruction} \\
        \midrule
        1 & put the ketchup in the wooden tray & Change of plan: put the ketchup on the plate instead of in the wooden tray. \\
        2 & Put the chocolate pudding on the plate. & Change of plan: put the chocolate pudding in the wooden tray instead of on the plate. \\
        3 & Put the chocolate pudding on the plate. & First put the chocolate pudding down on the table, turn on the stove and keep it on, and then continue putting the chocolate pudding on the plate. \\
        4 & put the chocolate pudding on the plate & Avoid the obstacle and continue putting the chocolate pudding on the plate. \\
        5 & Put the BBQ sauce in the wooden tray. & Change of plan: leave the BBQ sauce on the table and put only the chocolate pudding in the wooden tray. \\
        \bottomrule
    \end{tabularx}
    \caption{\textbf{Prompt shift tasks.} The instruction changes after a
    physical trigger without resetting the scene or robot.}
    \label{tab:prompt-change-tasks}
\end{table*}

The first three instruction changes occur during transport after the object has
been lifted and approached toward its original destination. Task 4 introduces
an obstacle and issues a reminder after the gripper-closing event. Task 5
redirects the target during the pre-grasp approach. The trigger for Task 4 uses
a positive closing command and three consecutive steps with less than 1\,mm
change in gripper width; this event alone does not certify a successful grasp. We note that the diagnostic is not intended to establish reliable low-level control, but to characterize VLMs’ local decision-making capabilities and limitations, motivating a mid-level action interface with continuous feedback and correction.

\subsection{Dynamic Environment Protocol}
\label{app:dynamic-environment-protocol}

\paragraph{Simulation stepping during planning.}
The dynamic protocol advances the environment according to elapsed wall-clock
time while the agent is planning or waiting for model inference. At the
20\,Hz control rate, the clock schedules one simulation step for each
accumulated 0.05\,s of non-action wall time. These steps keep the conveyor and
environment dynamics progressing while the robot receives a zero-motion
command. Planning latency therefore contributes to elapsed scene evolution
rather than providing a deliberately paused environment. Robot actions retain
their native execution schedule.

\paragraph{Observations and interventions.}
Agents receive the task instruction, their supported camera observations, and
robot measurements through their respective interfaces. They are not given
ground-truth target identities, scripted object trajectories, intervention
triggers, or evaluator state. Interactive interventions alter the physical
scene while retaining the instruction. Prompt Change retains the physical scene
and updates the instruction at the trigger. All methods for a given task use the
same task definition and intervention rule, although the trigger may be reached
at different times.

\paragraph{Instruction-update handling.}
Direct VLA policies discard queued actions associated with the old instruction
and use the revised text on subsequent inference. CAP-X interrupts the old
program at an action boundary and updates the generation goal. \MethodName{}
retains the same mission and interrupts only the active subgoal. The planner
generates a new plan for the revised instruction using the current observation,
the previous plan, accumulated memory, and recent execution evidence. The
physical environment is not reset, and cancellation and switching time remain
part of the episode budget.

\paragraph{Episode setup and budget.}
The online suites use a 600\,s wall-clock budget. Interactive and Prompt Change
use ten settling steps; the dynamic suites use 32 settling steps and an
additional 20{,}000-step horizon.

\paragraph{Success criteria.}
Task success is assessed using environment predicates rather than an agent's
declaration of completion. For tray placement in Prompt Change Tasks 2 and 5,
scoring requires the object origin to lie within the oriented tray region,
upward load-bearing support from the tray, and release from the gripper. The
implemented stove-interruption check records whether the stove was on before
final placement; it does not independently verify the requested intermediate
table placement or that the stove remains on at termination. For intervention
suites, task success and whether the intervention trigger was reached are
separate outcomes; episodes that fail before triggering an intervention remain
part of the task evaluation.

\section{Additional Ablation Study}
\label{app:ablation}

\begin{table}[t]
\centering
\scriptsize
\setlength{\tabcolsep}{5pt}
\renewcommand{\arraystretch}{1.08}
\begin{tabular}{lcccccc}
\toprule
\textbf{Budget}
& \textbf{Goal}
& \textbf{Spatial}
& \textbf{Object}
& \textbf{Avg. SR}
& \textbf{Time (s)}
& \textbf{Time score} \\
\midrule
300\,s
& 40.0\% & 70.0\% & 90.0\%
& \textbf{66.7\%}
& \textbf{159.8}
& \textbf{25.03} \\
450\,s
& 50.0\% & 70.0\% & 80.0\%
& \textbf{66.7\%}
& 210.0
& 19.05 \\
900\,s
& 40.0\% & 60.0\% & 50.0\%
& 50.0\%
& 320.0
& 9.37 \\
3,600\,s
& 50.0\% & 60.0\% & 70.0\%
& 60.0\%
& 296.4
& 12.14 \\
\bottomrule
\end{tabular}
\caption{\textbf{Execution-budget ablation.}
Average success does not increase monotonically with the maximum
execution budget. Time score uses the definition in Section~\ref{sec:experiments} (pp/min). The 300\,s setting matches the success of the
450\,s setting while requiring less wall time.}
\label{tab:budget_ablation}
\end{table}

\noindent\textbf{Effect of Execution Budget.}
Table~\ref{tab:budget_ablation} shows that both the 300\,s and default
450\,s settings achieve 66.7\% average success. The shorter budget
reduces mean wall time from 210.0\,s to 159.8\,s, a 23.9\% reduction,
and increases time-normalized success score from 19.05 to 25.03. Extending the
budget to 900\,s lowers average success to 50.0\%; the 3,600\,s
setting reaches 60.0\%. Thus, additional time alone does not
consistently improve task completion in this evaluation. An effective
harness must use new observations to decide when further action is
useful and when execution should stop.

\section{Error Analysis Details}
\label{app:error-analysis}

The analysis covers episodes behind the \MethodName{} rows of
Tables~\ref{tab:combined_results} is shown in the Figure below.

\begin{figure}[h]
    \centering
    \includegraphics[width=\textwidth]{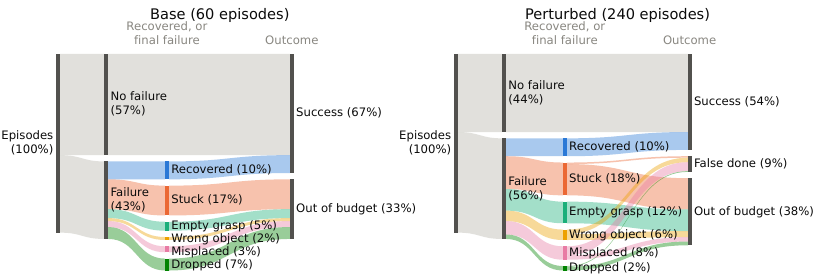}
    \caption{
    \textbf{Physical failure analysis} tracing \MethodName{} episodes through
    failures, recovery, and outcomes on unperturbed (left) and perturbed
    (right) LIBERO-PRO tasks. Each failed episode is assigned its last
    unrecovered failure; \emph{Recovered} episodes had failures and recovered
    from all of them. \emph{False done}: the model declared the task complete
    while its goal predicate was false. Band thickness is proportional to the
    share of episodes within each panel.
    }
    \label{fig:failure-flow}
\end{figure}

\end{document}